%% file: main.tex
\documentclass[nonacm,sigplan]{acmart}
\renewcommand\footnotetextcopyrightpermission[1]{}
\renewcommand{\shortauthors}{Amey Agrawal et al.}

\setcopyright{rightsretained}

\newboolean{publicversion}
\setboolean{publicversion}{true}

\input{packages}

\input{macros}

\date{}

\title{No Request Left Behind: Tackling Heterogeneity in Long-Context LLM Inference with \sysname}

\begin{document}

\author{Amey Agrawal$^{\text{2}}$\enskip Haoran Qiu$^{\text{1}}$\enskip Junda Chen$^{\text{3}}$\enskip Íñigo Goiri$^{\text{1}}$\enskip Chaojie Zhang$^{\text{1}}$\enskip Rayyan Shahid$^{\text{2}}$\enskip \\ Ramchandran Ramjee$^{\text{1}}$\enskip Alexey Tumanov$^{\text{2}}$\enskip Esha Choukse$^{\text{1}}$}
\affiliation{\vspace{1mm} $^{\text{1}}$Microsoft \enskip $^{\text{2}}$Georgia Institute of Technology  \enskip $^{\text{3}}$UC San Diego \country{}}

\input{0-abstract}

\maketitle

\input{0-outline}
\input{1-intro}
\input{2-background}
\input{3-insight}
\input{4-design}

\input{5-optimizations}

\input{6-eval}

\input{8-related}
\input{9-conc}

\bibliographystyle{ACM-Reference-Format}
\bibliography{all}

\appendix

\input{10-appendix}

\end{document}

%% file: packages.tex
\usepackage{amsmath}
\usepackage{enumitem}
\usepackage{xspace}
\usepackage{caption}
\usepackage{subcaption}
\usepackage{float}
\usepackage{cleveref}
\usepackage[small, compact]{titlesec}
\usepackage{ifthen}
\usepackage{comment}
\usepackage{graphicx}
\usepackage{xspace}
\usepackage{tcolorbox}
\usepackage{algorithm}
\usepackage{algpseudocode}
\usepackage[font=small,labelfont=bf]{caption}

\setlist[itemize]{noitemsep, topsep=1pt}
\setlist[enumerate]{noitemsep, topsep=1pt}

%% file: macros.tex
\newcommand{\sysname}{Medha\xspace}

\newcommand{\myx}{$\times$\xspace}

\newcommand{\llamaL}{Llama-3 70B\xspace}
\newcommand{\llamaS}{Llama-3 8B\xspace}

\newcommand{\ls}{LoongServe\xspace}

\newcommand{\sota}{state-of-the-art\xspace}

\newcommand{\ie}{\textit{i.e.}\xspace}
\newcommand{\eg}{\textit{e.g.}\xspace}

\ifthenelse{\boolean{publicversion}}{
    \newcommand{\grumbler}[3]{}
    \newcommand{\jm}[1]{}
    \newcommand{\ap}[1]{}
    \newcommand{\nk}[1]{}
    \newcommand{\rr}[1]{}
    \newcommand{\amey}[1]{}
    \newcommand{\alexey}[1]{}
    \newcommand{\inigo}[1]{}
    \newcommand{\esha}[1]{}
    \newcommand{\junda}[1]{}
    \newcommand{\hq}[1]{}

    \newcommand{\osdidel}[1]{}
}
{
    \newcommand{\grumbler}[3]{\xspace\textcolor{#3}{\bf #1: #2}}
    \newcommand{\esha}[1]{\grumbler{Esha}{#1}{brown}}
    \newcommand{\inigo}[1]{\grumbler{Inigo}{#1}{violet}}
    
    \newcommand{\alexey}[1]{\grumbler{AT}{#1}{blue}}
    \newcommand{\rr}[1]{\grumbler{Ram}{#1}{cyan}}
    \newcommand{\amey}[1]{\grumbler{Amey}{#1}{teal}}
    
    \newcommand{\junda}[1]{\grumbler{Junda}{#1}{violet}}
    \newcommand{\hq}[1]{\grumbler{HQ}{#1}{cyan}}

    \newcommand{\osdidel}[1]{\textcolor{red}{#1}}
}

\newcommand{\myparagraph}[1]{\vspace{\smallskipamount}\noindent\textbf{#1.\xspace}}

\definecolor{dgreen}{rgb}{0.0, 0.5, 0.0}

\definecolor{codegreen}{rgb}{0,0.6,0}

\newtcolorbox{takeawaybox}{
    colback=black!5,
    colframe=black,
    fonttitle=\bfseries,
    title=,
    boxrule=1pt,
    left=2mm,
    right=2mm,
    top=1mm,
    bottom=1mm,
}

%% file: 0-abstract.tex
\begin{abstract}
Deploying million-token Large Language Models (LLMs) is challenging because production workloads are highly heterogeneous, mixing short queries and long documents. This heterogeneity, combined with the quadratic complexity of attention, creates severe convoy effects where long-running requests stall short, interactive ones, degrading system responsiveness. We present \sysname, a serving system that eliminates these convoys by introducing fine-grained, preemptive scheduling to LLM inference.
    
\sysname makes preemption practical with a co-designed set of mechanisms -- including \textit{Adaptive Chunking} and \textit{Stream Pipeline Parallel}-- that overcome the perceived inefficiencies and scaling challenges of chunking. Additionally, we present a new parallelism strategy \textit{KV-Cache Parallelism} to reduce the decode latency and afford interactivity despite very long context.
These mechanisms are orchestrated by a \textit{Length-Aware Relative Slack (LARS)} scheduler, a deadline- and heterogeneity-aware  scheduling policy that prevents both the convoy effect and the starvation that plagues simpler policies. Under a heterogeneous workload, \sysname improves throughput by 5.7\myx while reducing median and 99th-percentile latency by 30\myx and 174\myx, respectively, compared to state-of-the-art non-preemptive systems.
\end{abstract}

%% file: 1-intro.tex
\section{Introduction}

Large language models with million-token context windows are transforming how we interact with information -- enabling large-scale document analysis, multi-hour video understanding \cite{gemini,llama4,qwen2.5}, and autonomous coding agents \cite{anthropic2025claudecode, google2025geminicli}.
However, deploying these models in production poses a significant challenge: real-world workloads combine both long and short requests. A single service must handle everything from 100-token chat messages to 10M-token document processing, often from the same users within the same session.

\input{figures/mnemosyne_banner_fig}

\textbf{Motivation.}
This mix of request lengths to the same model instance creates extreme computational heterogeneity due to the quadratic complexity of self-attention~\cite{attentionpaper}.
A 100K-token request is not 100\myx but approximately 10,000\myx more computationally expensive than a 1K-token request.
When these requests share the same serving infrastructure, we run into severe performance degradation due to: the \textit{convoy effect}~\cite{ostep,dino}, where long-running requests block shorter ones, resulting in poor system responsiveness.
As shown in \Cref{fig:banner}, even with just 5\% long requests in the workload, state-of-the-art systems like \ls~\cite{2024loongserve} experience 30\myx median latency increases and 174\myx tail latency degradation for short requests.

Context Parallelism (CP)~\cite{liu2023ring,brandon2023striped} has made it possible to distribute long-context processing across hundreds of GPUs, enabling effective training for long-context models.
\ls~\cite{2024loongserve} and Yang et al. \cite{yang2023cp}
adapted these techniques for inference by introducing elasticity to context parallelism, dynamically adjusting the degree of parallelism based on request length to improve resource efficiency.
However, these approaches fundamentally lack preemptability --- once a long request begins processing, it cannot be interrupted until completion.
This non-preemptive execution model inevitably results in convoy effects in heterogeneous workloads.
\Cref{fig:banner:hol} demonstrates this problem empirically: \ls exhibits severe latency spikes lasting for 100s of seconds whenever long requests block the system.
These spikes occur because arriving short requests must wait for the entire prefill computation of the long request to complete --- which can take several minutes.

The convoy effect is a well-studied problem in operating systems with a known solution: preemptive scheduling.
Chunked prefills \cite{agrawal2024taming} provide a natural mechanism for preemptable prefill computation.
However, applying preemption to LLM inference faces three barriers that have discouraged its adoption.
First, chunking long prefills are considered inefficient due to\textbf{\textit{ repeated KV-cache reads.}}
Second, batching decodes requests with chunked prefills for long requests results in \textbf{\textit{high decode latency}} that degrades user experience.
Finally, existing context parallelism techniques for long-context inference are \textbf{\textit{fundamentally incompatible with chunked execution}}. 

\input{figures/scheduling/convoy_effect_fig}

\textbf{Our work.}
\sysname makes preemptive long-context inference practical by systematically addressing each barrier.
We demonstrate that KV-cache read amplification is a non-issue for modern architectures -- chunks as small as 40 tokens achieve near-optimal efficiency due to high arithmetic intensity in grouped-query attention.
We introduce \textit{adaptive chunking}, which dynamically adjusts chunk sizes as the computational bottleneck shifts from MLP to attention operations, maintaining both high throughput and predictable latency.
We develop two parallelism strategies compatible with preemption: 
\textbf{\textit{Stream Pipeline Parallelism (SPP)}} accelerates prefills by pipelining chunks across stages, while \textbf{\textit{KV-Cache Parallelism (KVP)}} bounds decode latency by distributing attention computation.

To effectively leverage the preemptable prefills, we introduce Length-Aware Relative Slack (LARS), a deadline-aware  scheduling policy designed to explicitly tackle the heterogeneous nature of long-context inference. Unlike traditional policies that either cause convoy effects (First Come First Serve - FCFS) or starvation (Earliest Deadline First - EDF, Least Remaining Slack - LRS), LARS ensures both short and long requests meet their deadlines by pushing completions toward their SLO boundaries -- maximizing schedule robustness against unpredictable arrivals. Furthermore, we introduce a  dynamic batch packing algorithm that creates batches that maximally utilize GPU compute by co-locating complementary prefill chunks and decode requests while respecting strict time budgets.

\sysname unifies these techniques in a unified serving system that scales to multi-million token requests while maintaining high throughput and low latency. In summary, we make the following contributions in this paper:

\begin{itemize}[leftmargin=*,noitemsep,topsep=1pt]
    \item We identify and quantify the convoy effect in long-context inference arising due to request-length heterogeneity in current non-preemptive systems.

    \item We make chunked prefills viable for preemptive long-context inference by systematically addressing their perceived inefficiencies: introducing adaptive chunking to balance throughput and latency, and developing Stream Pipeline Parallelism (SPP) and KV-Cache Parallelism (KVP) as preemption-compatible parallelism strategies.

    \item We propose Length-Aware Relative Slack (LARS), a scheduling policy that prevents convoy effects and starvation by  accounting for workload heterogeneity and user SLOs.

    \item We implement \sysname with optimized kernels and scheduler design, demonstrating 5.7\myx higher throughput and up to 174\myx lower tail latency than state-of-the-art non-preemptive systems on real long-context workloads.
\end{itemize}

%% file: figures/mnemosyne_banner_fig.tex
\begin{figure}[t!]
    \begin{subfigure}{0.48\linewidth}    \includegraphics[width=\linewidth]{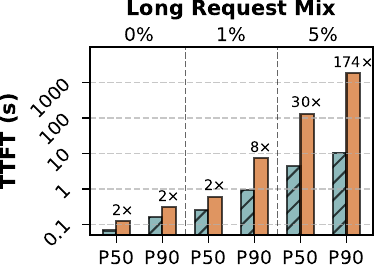}
        \vspace{0em}
        \caption{
            TTFT distribution where state-of-the-art baseline shows 30-174\myx{} higher latency due to lack of preemptive scheduling.
        }
        \label{fig:banner:bars}
    \end{subfigure}
    \begin{subfigure}{0.48\linewidth}          
    \includegraphics[width=0.88\linewidth]{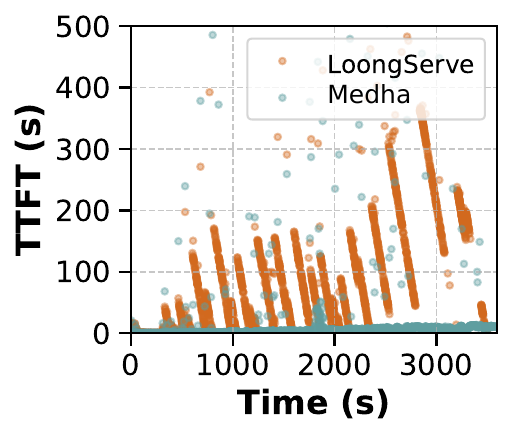}
        \caption{
            TTFT over time showing severe convoy effect for \ls{} where short requests get stuck behind long requests.
        }
        \label{fig:banner:hol}
    \end{subfigure}
    \caption{
        Impact of long-context requests on TTFT for \llamaS{} inference using 16 A100 GPUs with \ls{}~\cite{2024loongserve} and \sysname{} at 0.75 QPS.
    }
    \label{fig:banner}    
\end{figure}

%% file: figures/scheduling/convoy_effect_fig.tex
\begin{figure}[t!]
    \centering
    \includegraphics[width=\linewidth]{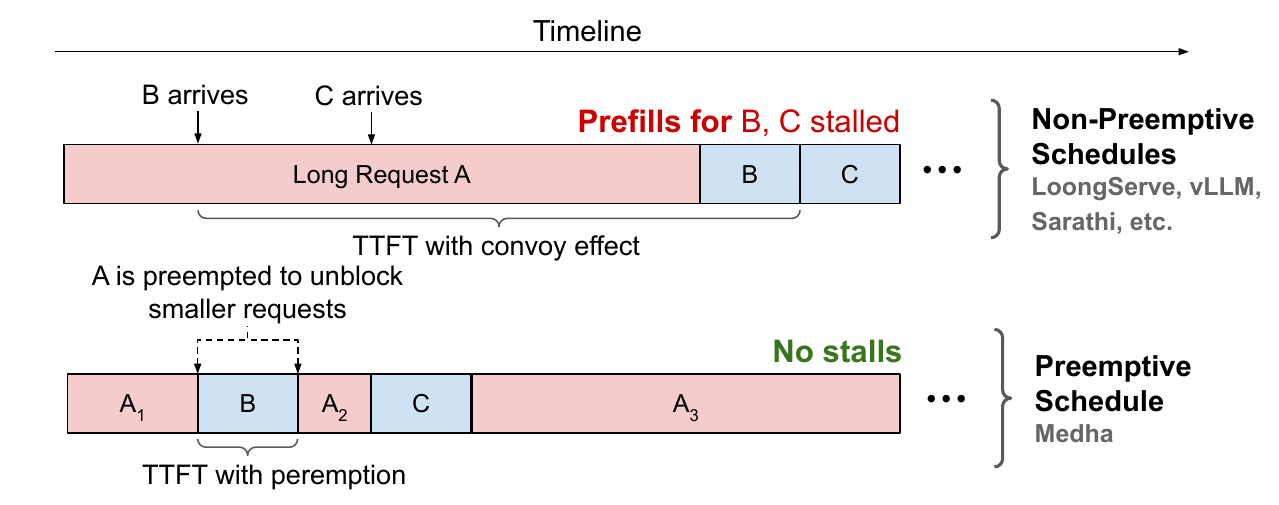}
    \caption{Impact of preemption on convoy effect. Non-preemptive scheduling (top) blocks short requests B and C behind long request A, causing deadline violations. Preemptive scheduling (bottom) interleaves execution through chunking, eliminating convoy effect while maintaining throughput.}
    \label{fig:convoy}
\end{figure}

%% file: 2-background.tex
\section{Motivation: The Case for Preemptive Long Context LLM Inference}
\label{sec:challenges}

In this section, we analyze how million-token LLM inferences create extreme computational heterogeneity due to quadratic attention complexity, causing convoy effects where long requests block shorter ones, motivating our preemptive inference approach.

\subsection{Background: LLM Inference Characteristics}
\label{sec:background-condensed}

Auto-regressive LLM inference comprises two fundamentally different phases with distinct performance characteristics~\cite{sarathi2023,patel2023splitwise,distserve2024}. The \textbf{prefill phase} processes the entire prompt through a single forward pass to construct Key-Value (KV) cache and generate the first token. This phase is compute-bound, with performance measured by Time-to-First-Token (TTFT). The subsequent \textbf{decode phase} generates tokens autoregressively, one at a time, and is memory-bandwidth-bound. Its performance is measured by Time-Per-Output-Token (TPOT) or Time-Between-Tokens (TBT) \cite{etalon}.

Contemporary serving systems employ two primary parallelism strategies.  \textbf{Tensor Parallelism (TP)}~\cite{megatron} partitions each model layer across multiple devices. This reduces per-device memory requirements and can improve latency. However, TP requires high communication bandwidth, limiting it to single servers with fast interconnects like NVLink. \textbf{Pipeline Parallelism (PP)}~\cite{gpipe, orca, sarathi2023} distributes complete layers sequentially across devices. While this reduces memory pressure per device and can improve throughput, it provides no latency benefit for individual requests due to its sequential execution model.

\subsection{Context Length Scaling Limits of Conventional Parallelism Techniques}
\label{sec:limits-conventional}

\myparagraph{Memory Constraint} In the prefill phase, since all the input tokens are processed concurrently, the activation memory required for prefill computation increases linearly with the context length. While tensor parallelism can distribute this load across devices, it cannot be scaled beyond a single node due to communication overhead. %

\myparagraph{Latency Constraints} The quadratic complexity of attention operation becomes a major challenge for interactive workloads as sequence length grows. For instance, to process 1M tokens with \llamaL we require a total of 2.8 ExaFLOPs. On an H100 GPU, even at full utilization, this computation would require at least 48 minutes to execute. To perform this computation in a reasonable time, the attention computation needs to be parallelized across a large number of GPUs. However, neither tensor nor pipeline parallelism provides a viable solution. As discussed previously, TP does not scale beyond a single node (8 GPUs) due to communication overhead \cite{narayanan2021efficient, sarathi2023}, while PP can scale to a large number of GPUs, it does not provide any latency advantage.

\begin{takeawaybox}
\textit{\textbf{Takeaway:}} Standard parallelization techniques like TP or PP fail at million-token contexts due to memory limits and quadratic attention costs that lead to high latency.
\end{takeawaybox}

\subsection{Long-Context System Scaling with Context Parallelism}
\label{sec:ring-attention-scaling}

To overcome the memory and latency limitations of conventional parallelism, Liu et al. introduced \textbf{Context Parallelism (CP)} \cite{liu2023ring,brandon2023striped} for long-context \textit{training}. In CP, the input sequence is partitioned across multiple GPUs to alleviate activation memory pressure. By overlapping KV block communication between GPUs with computation, CP enables efficient scaling to hundreds of devices. This approach has been widely adopted in long-context training systems.

However, context parallelism's design is fundamentally misaligned with the demands of inference serving. To achieve efficient overlap of communication and computation in CP, each GPU must process a sufficiently large sequence partition (e.g., 24.5K tokens on A100 with InfiniBand \cite{liu2023ring}). This creates a critical latency-throughput tradeoff --- a system configured with high parallelism degrees for low-latency serving of long context requests suffers from severe underutilization when serving short requests. Conversely, a system configured for short requests cannot achieve acceptable latency for long ones.

\subsection{Adapting Context Parallelism for Inference}
\label{sec:adapting-ring-attention}

To address the rigid resource allocation in CP, the state-of-the-art system \ls \cite{2024loongserve} adapts it for inference by introducing two key mechanisms. First, it proposes an elastic version of context parallelism, where the degree of parallelism is dynamically adjusted to match the workload --- allocating more resources to accelerate long, compute-intensive prefills while using fewer for short requests, thereby improving efficiency.

Second, because CP is ineffective for decode phases, the system must adopt the prefill-decode disaggregation paradigm \cite{patel2023splitwise, distserve2024}. In this model, the prefill and decode phases are handled by separate, isolated groups of GPUs. After a request's prefill is complete, its KV cache is migrated from the prefill pool to a different group of GPUs dedicated to the less resource-intensive decode phase. Furthermore, since the relative prefill to decode load ratio in the system dynamically changes based on the input requests pattern \cite{mitra2025beyond}, \ls adopts an elastic approach where the number of GPUs in the prefill/decode pool is dynamically adjusted to match the workload. This elastic, disaggregated architecture represents the current state-of-the-art approach for long-context inference --- achieving 3-5\myx \cite{2024loongserve} lower latency than prior systems like vLLM \cite{vllmsosp}, DistServe \cite{distserve2024}, and Sarathi-Serve \cite{agrawal2024taming}.

\subsection{The Convoy Effect from Extreme Workload Heterogeneity}
\label{sec:convoy-effect-heterogeneity}

\hq{why loongserve doesn't have such issue? I guess here we also need to answer the question, why we don't want to have separate clusters to serve long vs. short requests}

\esha{we do show LS to have this problem. But we can add something to the effect of - having pools creates fragmentation, there is huge variance in context length within a session - leading to many KV transfers, with multi-million context length, how many pools would you even have etc}
The key challenge in serving long-context models stems from the extreme workload heterogeneity created by the quadratic complexity of self-attention. Because the required FLOPs for the attention mechanism scale with the square of the sequence length ($N^2$), the difference in processing time between requests grows superlinearly. For instance, a 100K-token request is not 100\myx but roughly 10,000\myx more computationally expensive than a 1K-token request. This extreme heterogeneity, lead to a classic systems challenge known as the \textit{\textbf{convoy effect}}~\cite{ostep,dino}. When the system processes a long prefill, all subsequent short requests behind it in the queue are stalled. As shown in \Cref{fig:banner:bars}, this leads to a complete collapse in system performance, increasing median TTFT by 30\myx and tail latency by 174\myx with just 5\% long requests in the workload.

\begin{takeawaybox}
    \textit{\textbf{Takeaway:}} The quadratic cost of attention creates extreme workload heterogeneity, leading to the \textbf{\textit{convoy effect}}, where long requests block short ones.
\end{takeawaybox}

\subsection{The Path to Preemption: Fine-Grained Chunking}
\label{sec:path-to-preemption}

The convoy effect is a widely studied problem in operating systems --- resolving this issue requires a shift from non-preemptive to preemptive scheduling \cite{ostep,dino}. To apply this principle to LLM serving, the long, atomic prefill operation must be broken down into smaller, interruptible units of work. Chunking the input prompt \cite{agrawal2024taming} achieves exactly this, creating scheduling opportunities to interleave short requests with long ones.

However, naive chunking is widely considered impractical for long contexts due to three prohibitive systems challenges, which this paper systematically resolves:

\myparagraph{KV-Cache Read Amplification} In a standard, non-chunked prefill, the KV cache is read once. With chunking, however, the processing of each subsequent chunk requires re-reading the entire KV cache generated by all previous chunks from GPU memory. This transforms the memory access pattern from being linear with the sequence length to being quadratic. Because memory bandwidth is a critical and often limited resource, this quadratic increase in data movement has led to the widespread belief that chunking is fundamentally inefficient and unscalable for long-context serving \cite{distserve2024,sarathi2023}.

\myparagraph{Latency Interference} Piggybacking prefill chunks \cite{agrawal2024taming} onto decode batches is a standard technique to improve GPU utilization and reduce tail latency by co-executing compute-bound prefill operations with memory-bound decode operations. However, with long contexts, the compute cost of successive prefill chunks grows quadratically as the context lengthens. Late-stage chunks become so computationally intensive that they stall latency-sensitive decodes, making it infeasible to batch prefill chunks with latency-sensitive decodes --- for instance, computing the a prefill chunk for a 1M context request with \llamaS on 8 H100s using chunk size of 512 results in decode latency of $\sim 250$ms, almost an order of magnitude higher the typical production SLOs.

\myparagraph{Lack of a Preemption-Friendly Scaling Strategy} Adopting chunking means abandoning the only proven technique for scaling prefill latency -- context parallelism. While CP operates by splitting the sequence in a special dimension across different devices, chunked prefill unrolls the prefill computation in a temporal dimension by processing each prefill chunk sequentially. There is no trivial solution to combine the two approaches. The conventional alternatives are insufficient for serving long contexts. This creates the need for entirely new parallelism strategies designed to be both scalable and preemptive.

%% file: 3-insight.tex
\section{Enabling Efficient Preemptable Prefills} \label{sec:mechanisms}
To enable preemptive execution using chunked prefills we need to overcome three perceived barriers: the belief that chunking causes prohibitive KV-cache read amplification, concerns about latency interference between chunked operations, and the lack of preemption-friendly parallelism strategies. In this section, we present the insights that allow \sysname to systematically address these challenges.

\input{figures/chunked_prefills/chunking_overhead_attn_only_fig}

\input{tables/notation}

\subsection{Debunking KV-Cache Read Amplification \\Inefficiency Myth} \label{sec:kv-read-amplification}

Conventional wisdom maintains that chunked attention is inherently inefficient due to KV-cache read amplification---the repeated reading of cached keys and values across chunks. We analyze the attention computation from first principles and demonstrate this assumption is incorrect for modern model architectures.

\myparagraph{Arithmetic Intensity Analysis} 
Modern GPU architectures feature independent compute and memory subsystems that operate concurrently in a pipelined fashion. Performance is determined by whichever subsystem becomes saturated first. When the compute subsystem is fully utilized, additional memory operations execute in parallel \textit{``for free''} without impacting the device throughput.

The key metric determining this behavior is arithmetic intensity---the ratio of compute operations to memory accesses. High arithmetic intensity indicates sufficient computation per byte of data to keep compute units busy while memory transfers complete in parallel. For chunked attention, this relationship is governed by:

\begin{equation}
I^{i}_{cp}(n, c) \simeq \frac{4ic^2dh_{q}}{4icdh_{kv}} = c \frac{h_{q}}{h_{kv}}
\label{eq:chunkedprefillai}
\end{equation}

The critical insight is that arithmetic intensity for chunked attention depends solely on chunk size, not total context length. Each chunk processes $c$ tokens, requiring reads of the full KV cache but performing a fixed number of operations per token. Furthermore, contemporary LLMs employ Grouped-Query Attention architectures where multiple query heads share KV heads (e.g., 8$\times$ in \llamaL), resulting in high arithmetic intensity such that even small chunks can saturate GPU compute.

\myparagraph{Empirical Validation} 
On H100 GPUs running \llamaL, chunks of just ~40 tokens can saturate GPU compute. We find that for 1M-token contexts, 32-token chunks incur merely 11\% overhead relative to 2048-token chunks as shown in \Cref{fig:chunkedprefill:overheadattnonly}. Note that, unlike older attention implementations used prior analysis \cite{sarathi2023} --- where the chunked prefill was shown to be inefficient for long contexts --- modern attention kernels like FlashInfer and FlashAttention-2~\cite{flashinfer, flashattention2} parallelize over both query and KV dimensions, which helps in materializing the theoretical performance potential of the chunked prefill.

\input{figures/experiments/prefill_decode_tradeoff/prefill_decode_tradeoff_combined_fig}

\subsection{Managing Interference with Adaptive Chunking}
\label{sec:adaptive-chunking}

Having established that chunking is computationally efficient, we now address the prefill-decode latency interference in mixed batches.

\myparagraph{The Throughput-Latency Tradeoff of Chunking}
Chunked prefills face a fundamental throughput-latency tradeoff. To maximize system throughput, a scheduler must use large chunks to process prefills efficiently; however, to guarantee low decode latency for co-batched requests, it must use small chunks. This tradeoff would be trivially resolved if we could execute small chunks with minimal overhead. As shown in \cref{sec:kv-read-amplification}, even chunks as small as 40 tokens are enough to saturate \textit{attention} computation --- however, the challenge arises because different operations show varying performance characteristics. The MLP component has a significantly lower arithmetic intensity than attention --- $c$ as opposed to $c \frac{h_{q}}{h_{kv}}$ for attention, and is thus more sensitive to chunk size. Moreover, there are fixed per-chunk overheads like kernel launches that are not amortized by chunking. As shown in \Cref{fig:prefilldecode:tradeoff}, this conflict forces an undesirable choice between high throughput and low decode latency.

\myparagraph{Resolving the Tradeoff with Adaptive Chunking}
To resolve the throughput-latency tradeoff, our approach is based on the key insight that a long prefill's computational bottleneck is not static, but \textit{shifts} as the prefill progresses. Prefill computation is initially dominated by MLP layers, which require large chunks to run efficiently and achieve high throughput. As the KV cache grows, the quadratic cost of the attention operation becomes the overwhelming dominant cost. At this stage, a switch to smaller chunks is possible. While smaller chunks make the MLP computation slightly less efficient, this is an acceptable trade-off because the performance hit is negligible compared to the now-dominant cost of attention. Based on this insight, \sysname implements an \textbf{Adaptive Chunking} policy. The policy begins a prefill with large chunks and dynamically shrinks them as the bottleneck shifts, thereby maintaining a predictably low iteration time. This adaptive strategy resolves the tradeoff faced with static chunking, achieving both high prefill throughput and low decode latency, as shown in \Cref{fig:prefilldecode}.

\subsection{Scalable Parallelism for Preemptive Inference}
\label{sec:parallelism}

To achieve interactive latency for million-token requests, preemptive chunking must be combined with a scalable, multi-node parallelism strategy. As existing approaches are incompatible with our preemptive model, \sysname introduces two novel techniques: Stream Pipeline Parallelism and KV-Cache Parallelism.

\input{figures/parallel_strats/spp_sched_combined_fig}

\myparagraph{Accelerating Prefill Computation}
We leverage an overlooked opportunity to accelerate prefill computation: while prior works process each chunk sequentially through all model layers, we observe that the chunks of a single request can be processed concurrently across pipeline stages.

In chunked prefills, chunk $i+1$ requires the KV cache from chunk $i$, but critically, it does \emph{not} need chunk $i$'s final model output. This means chunk $i+1$ can begin processing as soon as chunk $i$ completes the first pipeline stage; it does not need to wait for chunk $i$ to finish all pipeline stages.

Traditional approaches that combine chunking with pipeline parallelism treat each chunk like a separate request, processing them sequentially through the entire pipeline. This leaves pipeline stages underutilized. For example, when chunk $i$ moves from Stage 1 to Stage 2, Stage 1 sits idle. To fill these "pipeline bubbles," different requests are interleaved across stages, a technique known as micro-batching.

In contrast, \sysname introduces \textbf{Stream Pipeline Parallelism (SPP)}, which exploits the unique data dependency structure of chunked prefills. It schedules chunk $i+1$ to start Stage 1 immediately when chunk $i$ advances to Stage 2 \Cref{fig:sppcombined:sched}. This allows layers across all pipeline stages to operate concurrently on different chunks of the same long prefill, reducing the critical path of the prefill by the pipeline depth.

This chunk-level pipelining scales effectively. As shown in \Cref{fig:sppcombined:scaling}, Stream Pipeline Parallelism scales nearly linearly with the number of stages, enabling inference on multi-million-token requests using hundreds of GPUs. Please refer to \Cref{appendix:scaling} for scaling results up to 10M tokens. Compared to context parallelism, stream pipeline parallelism is significantly faster, achieving 1.64\myx lower latency on a one-million-token prefill and reducing TTFT over 128 H100s.

\input{figures/experiments/cp_prefill_scaling/cp_prefill_scaling_timeline_fig}

\vspace{-0.2em}
\myparagraph{Bounding Decode Latency}
While SPP addresses prefill latency, the decode phase presents its own scaling challenge. For long contexts, decode latency grows linearly with the sequence length, leading to high TPOT and a poor user experience. Furthermore, in mixed batches, even the smallest efficient prefill chunk (as determined by Adaptive Chunking) can still significantly slow down the decode operation when operating with very long contexts or larger models, creating a need for an additional mechanism to control iteration time.

\sysname introduces \textbf{KV-Cache Parallelism (KVP)} as a unified mechanism to address both challenges by parallelizing the KV cache reads across multiple devices along the sequence dimension. During any computation step (either a decode token or a prefill chunk), the query is replicated to each device, which computes a partial attention output in parallel using its local shard. These partial outputs are then combined using online-softmax.

KVP provides two critical, complementary benefits. First, for the decode phase, it places an upper bound on TPOT by ensuring the decode time can be capped for long requests, as shown in \Cref{fig:kvpresults:latency}. Second, for the prefill phase, KVP offers a new lever to manage latency interference that is complementary to Adaptive Chunking. It allows the scheduler to use larger, more throughput-efficient chunks while still meeting decode SLOs by parallelizing the chunk's internal attention computation, improving the overall TTFT-TPOT tradeoff as shown in \Cref{fig:kvpresults:tradeoff}. \sysname uses a \textit{progressive scaling} strategy, dynamically adding KVP workers as the context grows to maintain a near-consistent iteration latency. We provide additional KVP results in the appendix \Cref{appendix:scaling}.

\input{figures/parallel_strats/kvp_results_combined}

\vspace{-0.5em}
\begin{takeawaybox}
\textbf{\textit{Takeaway:}} When paired with mechanisms for scaling computation, chunked prefills provide a viable foundation for preemptive long-context inference.
\end{takeawaybox}

%% file: figures/chunked_prefills/chunking_overhead_attn_only_fig.tex
\begin{figure}[t!]
    \centering
    \begin{subfigure}[b]{0.49\linewidth}
        \includegraphics[width=\linewidth]{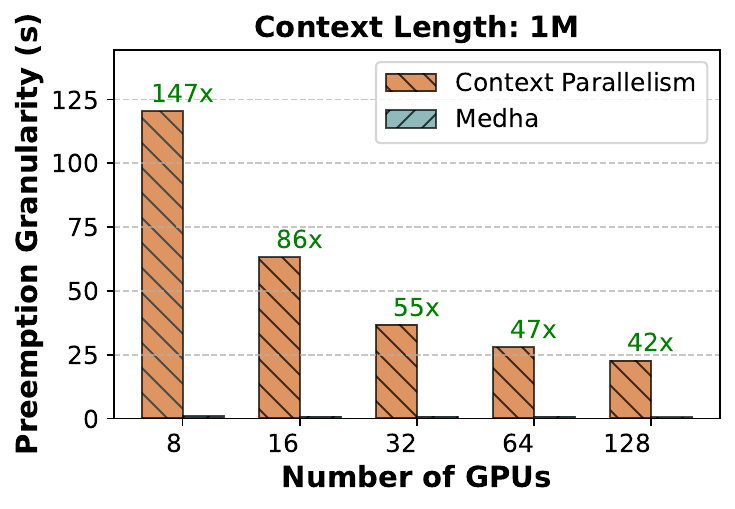}
        \caption{Preemption granularity enabled on 1M token sequences prefill with Llama-3 8B.}
        
        \label{fig:sppsched:preemptiongran}
    \end{subfigure}
    \begin{subfigure}[b]{0.49\linewidth}   
    \includegraphics[width=1\linewidth]{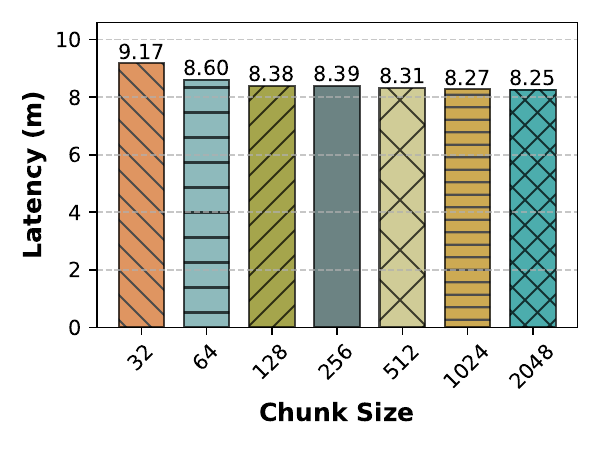}
    \vspace{-0.25in}
    \caption{Self-Attention computation time with chunked prefill for 1M tokens with \llamaL{} using 8 H100s.
    }
    \label{fig:chunkedprefill:overheadattnonly}    
    \end{subfigure}
    \caption{Efficacy of chunked prefill for long-context inference.}
\end{figure}

%% file: tables/notation.tex
\begin{table}[t]
\setlength{\abovecaptionskip}{2pt}
\setlength{\belowcaptionskip}{0pt}
    \centering
    \tiny{
    \footnotesize
    \caption{Definitions of notations in equations.}
    \label{tab:notation}
    \begin{tabular}{cc}
        \toprule
        \textbf{Notation} & \textbf{Definition} \\
        \midrule
        $n$ & number of tokens \\
        $h_q$ or $h_{kv}$ & number of query or key-value heads \\
        $d$ & attention head dimension \\
        $p_{j}$ & parallelism degree for strategy \emph{j}. \eg $p_{tp}$ for TP \\
        $I$ & arithmetic intensity \\
        $c$ & chunk size \\
        \bottomrule
    \end{tabular}
    }
\end{table}

%% file: figures/experiments/prefill_decode_tradeoff/prefill_decode_tradeoff_combined_fig.tex
\begin{figure}
    \centering
    \begin{subfigure}[b]{0.49\linewidth}        
        \centering\includegraphics[width=\linewidth]{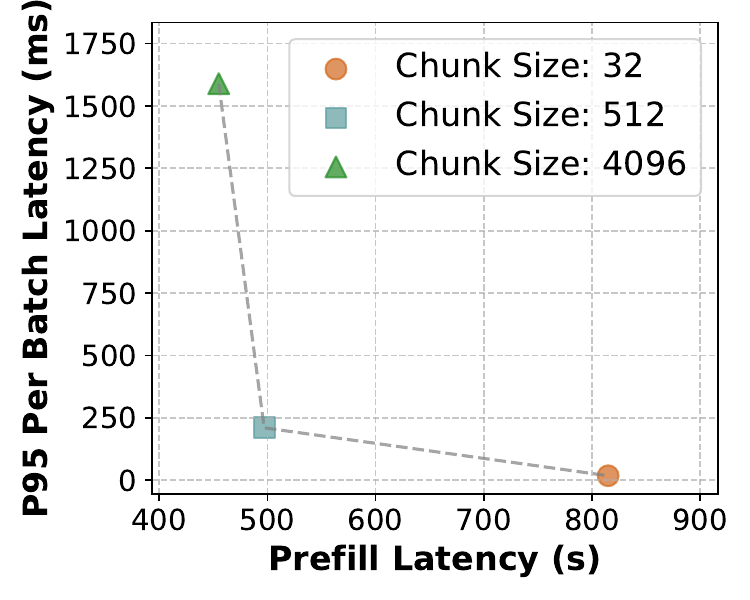}
        \caption{
            Static chunk sizes.
        }
        \label{fig:prefilldecode:tradeoff}
    \end{subfigure}
    \begin{subfigure}[b]{0.49\linewidth}
        \centering\includegraphics[width=\linewidth]{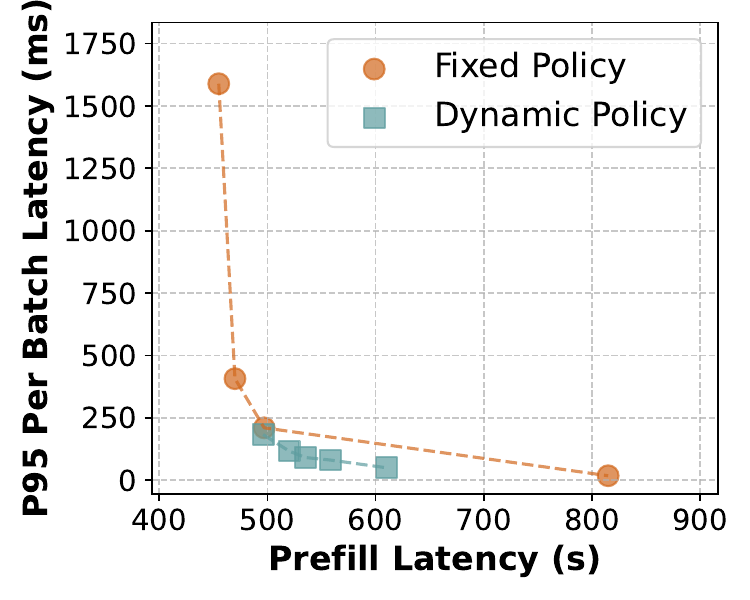}
        \caption{
            Adaptive chunk size.
        }
        \label{fig:prefilldecode:dynamictradeoff}
    \end{subfigure}
    \caption{
        Pareto frontiers of prefill/decode latencies in mixed batching with chunked prefills:
        (a) Static sizes have a trade-off between prefill and decode latencies.
        (b) Adaptive chunking starts with larger chunks, gradually reducing size to keep batch latencies consistent, achieving better prefill efficiency and low decode latency.
    }
    \label{fig:prefilldecode}
\end{figure}

%% file: figures/parallel_strats/spp_sched_combined_fig.tex
\begin{figure}
    \centering
    \begin{subfigure}[b]{0.45\linewidth}
    \centering
    \includegraphics[width=0.85\linewidth]{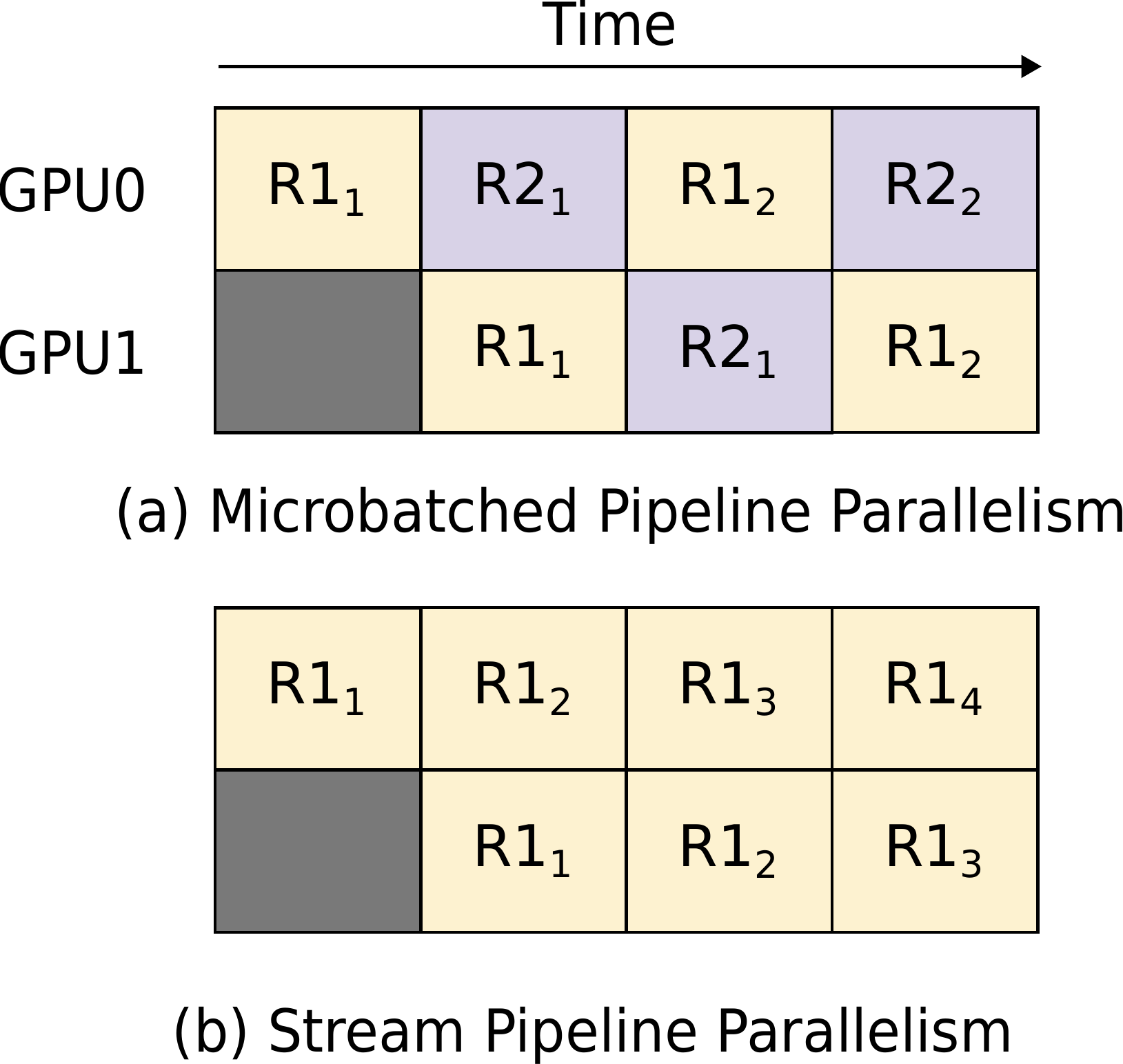}
    \caption{Contrasting PP strategies for prefill processing.}
    \label{fig:sppcombined:sched}    
    \end{subfigure}
    \begin{subfigure}[b]{0.45\linewidth}
    \centering
    \includegraphics[width=\linewidth]{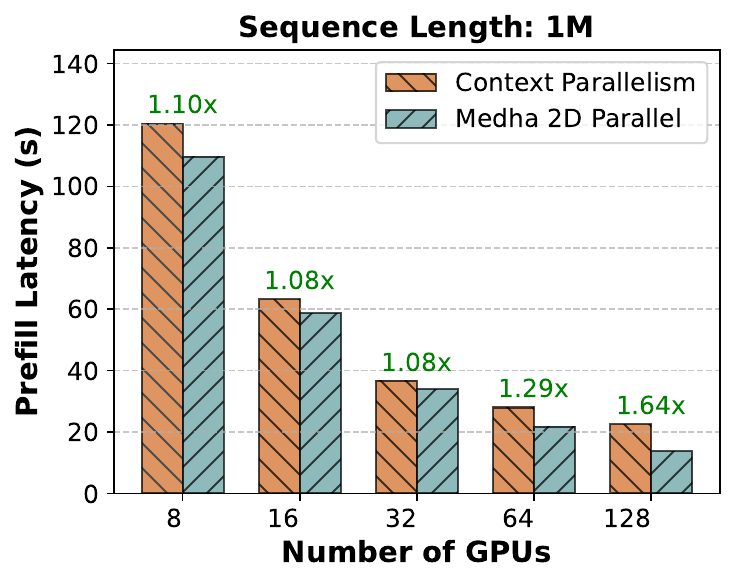}
    \caption{Performance comparison of Context Parallelism vs SPP+TP.}
    \label{fig:sppcombined:scaling}    
    \end{subfigure}
    \caption{Microbatched pipeline parallelism interleaves micro-batches composed of prefills from different requests ($R1$, $R2$) to improve throughput. SPP on the other hand, overlaps chunks of the same request ($R1_1$, $R2_2$) across stages to accelerate prefill processing. SPP achieves better scaling compared to CP due to lower communication overhead, resulting in up to 1.64\myx lower prefill latency for 1M context processing for \llamaS with H100s.}
    \label{fig:sppcombined}    
\end{figure}

%% file: figures/experiments/cp_prefill_scaling/cp_prefill_scaling_timeline_fig.tex
\begin{figure}[t!]
    \centering
    \begin{subfigure}[b]{0.4\linewidth}
    \centering    
    \includegraphics[width=0.9\textwidth]{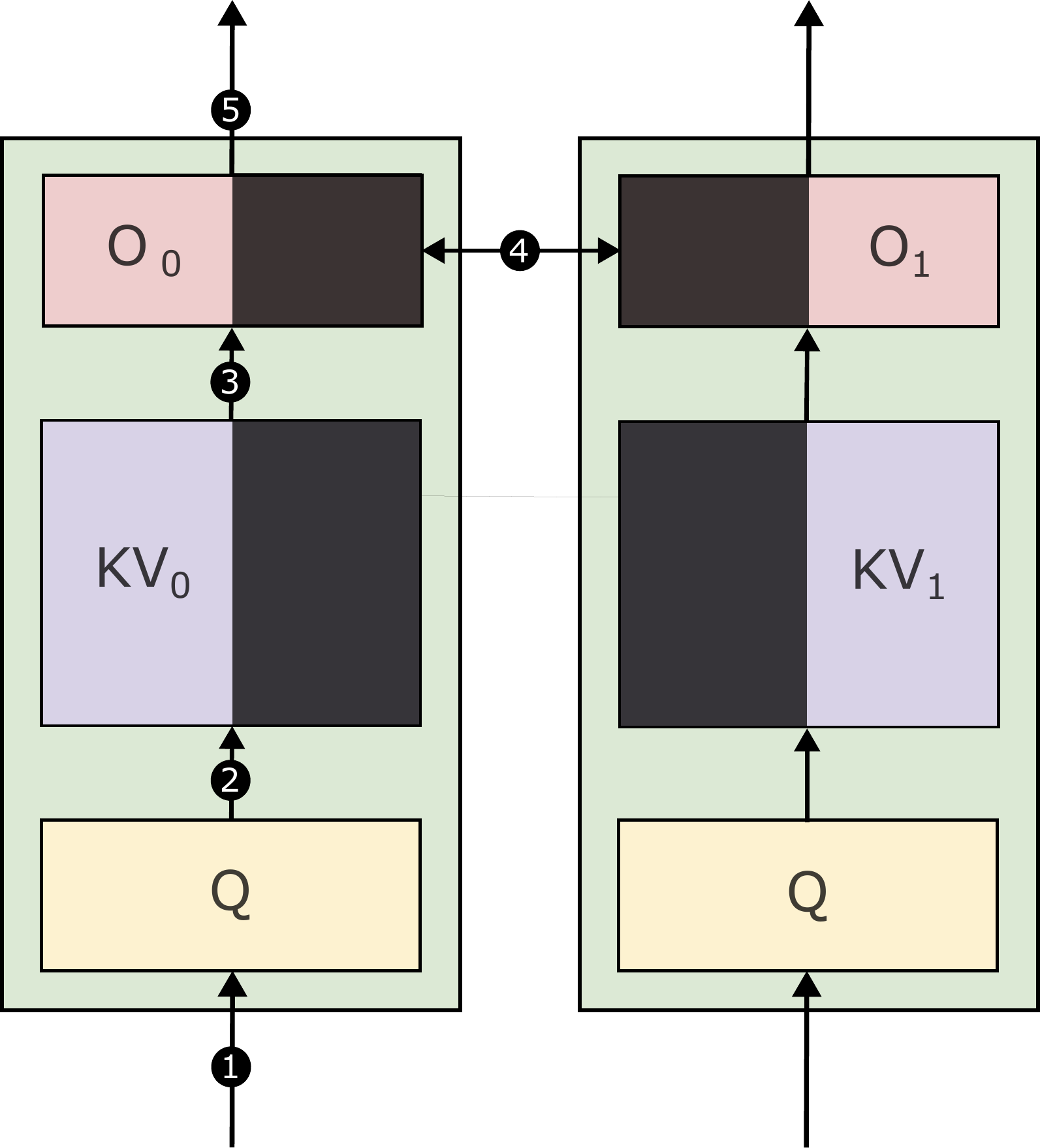}
    \caption{Sharding schema in KVP.}
    \label{fig:kvp:schema}
    \end{subfigure}
    \hfill  %
    \begin{subfigure}[b]{0.57\linewidth}
    \centering
    \includegraphics[width=\textwidth]{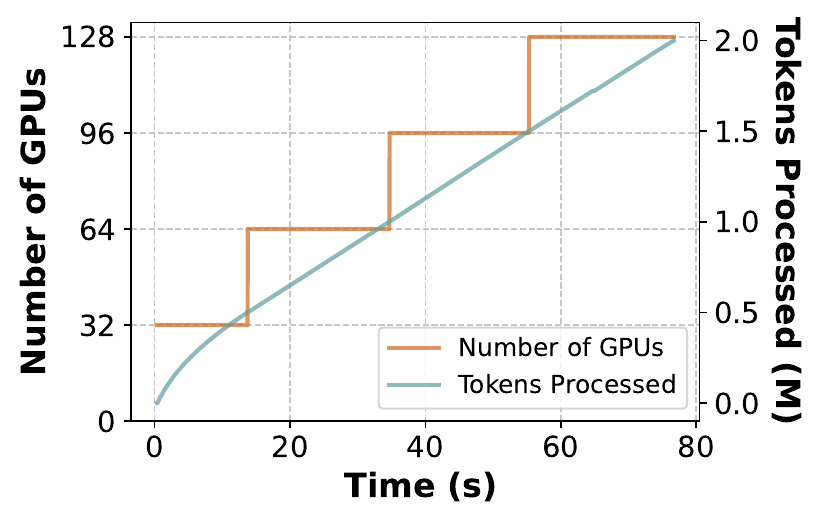}
    \caption{KVP Execution Timeline.}
    \label{fig:kvp:scaling}
    \end{subfigure}
    
    \caption{KV Parallelism distributes the KV-cache state across GPUs to minimize the latency of long requests. During prefill, KVP dynamically scales resources as the KV produced grows to maintain consistent iteration times irrespective of context length.  While processing 2M tokens with \llamaS using $p_{tp}=8$, $p_{kvp}=4$, and $p_{spp}=4$.
    \sysname{} starts with a single KVP worker group (4 servers) and progressively scales up to 16 servers.
    }
    \label{fig:kvp}
\end{figure}

%% file: figures/parallel_strats/kvp_results_combined.tex
\begin{figure}
\centering
    \begin{subfigure}[b]{0.49\linewidth}
        \includegraphics[width=\linewidth]{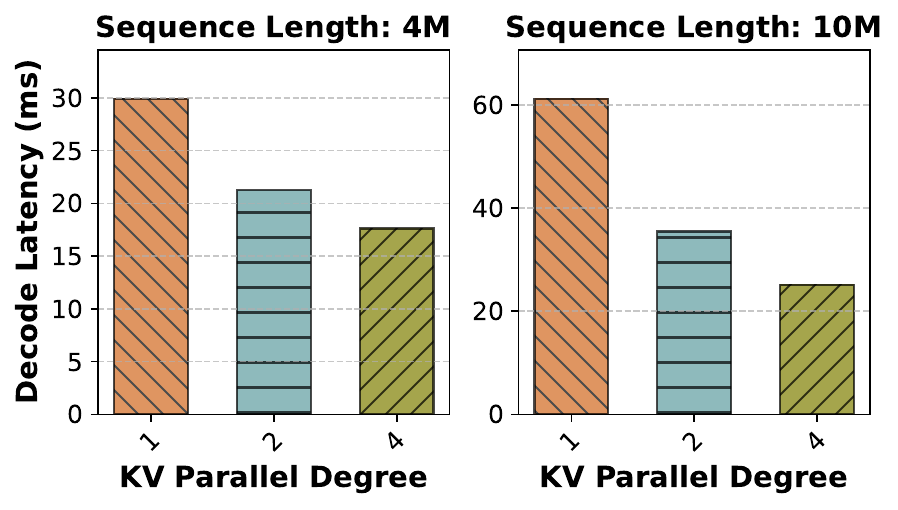}
        \caption{\llamaS with $p_{spp}=4$.}
        \label{fig:kvpresults:latency}
    \end{subfigure}
    \hfill
    \begin{subfigure}[b]{0.49\linewidth}
        \centering\includegraphics[width=0.8\linewidth]{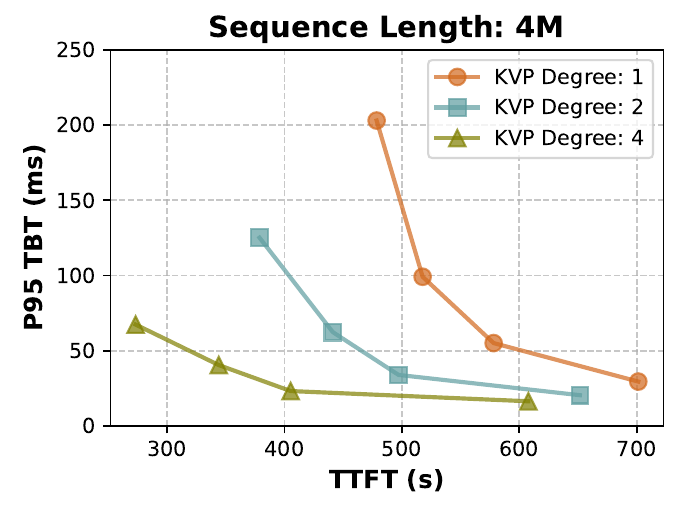}
        \caption{TTFT vs. TPOT trade-off space.}
        \label{fig:kvpresults:tradeoff}
    \end{subfigure}
    \caption{
     (a)TPOT reduction with KVP in \sysname in decode-only batches. For 10M context length decodes for \llamaS, $p_{kvp} = 2$ results in almost 40\% reduction in latency, allowing decode at the rate of $\sim$30 tokens per second. (b) KVP enables co-batching of larger prefill chunks with decode requests by parallelizing attention computation, reducing prefill-decode latency interference and providing a richer tradeoff space.
    }
    \label{fig:kvpresults}
    \end{figure}

%% file: 4-design.tex
\section{Scheduling Policies for Preemptive Inference}
\label{sec:policies}

The mechanisms presented in \Cref{sec:mechanisms} provide the necessary tools for preemptive inference for long-context requests. However, these are not sufficient on their own. To effectively navigate the throughput-latency tradeoff and resolve convoy effects with a mix of long and short requests, a robust scheduling policy is required. \Cref{fig:hld} illustrates how \sysname orchestrates these components: the Replica Controller implements our scheduling policies through a slack-aware Batch Scheduler that maintains request priorities using LARS, and a Batch Packer that constructs optimal batches guided by runtime predictions. These batches are then dispatched to the 3D Parallel Execution Engine, which leverages our novel combination of KVP, SPP, and TP to execute them efficiently. This section details the scheduling policies that drive these components—how \sysname prioritizes requests to prevent convoy effects (\Cref{sec:prioritization}), co-locates complementary prefills to improve throughput (\Cref{sec:multi-prefill-batching}), and packs batches to meet strict SLO requirements (\Cref{sec:batch-packing}).

\subsection{An SLO-Aware Prioritization Policy}
\label{sec:prioritization}

\input{figures/parallel_strats/3dp_fig}

In this section, we develop an online scheduling policy that prevents convoy effects while avoiding starvation, operating with sub-millisecond overhead. We analyze why widely used policies for LLM inference --- FCFS, EDF, and LRS fail under extreme heterogeneity, then present the approach adopted by \sysname{}: Length-Aware Relative Slack (LARS).

\myparagraph{Problem Formulation}
We consider a stream of requests $\mathcal{R} = \{r_1, r_2, ...\}$ where each request $r_i$ is characterized by: arrival time $a_i$, total work requirement $w_i^{\text{total}}$ (total computation time), and deadline $d_i$ relative to its arrival. At any time $t$, a request has remaining work $w_i(t)$ where $w_i(a_i) = w_i^{\text{total}}$. The scheduler must assign each request to time slots on available GPUs to maximize goodput -- the fraction of requests meeting their deadlines, without introducing any systematic bias towards long or short requests.

\myparagraph{Illustrative Scenario}
To understand how different policies handle heterogeneous workloads, consider a simple instance with three requests. Let request $r_L$ have $w_L^{\text{total}} = 10$ seconds with deadline $d_L = 16$ seconds. Let requests $r_S^1, r_S^2$ each have $w_S^{\text{total}} = 0.5$ seconds with deadlines $d_S = 1$ second. Request $r_L$ arrives at time 0; $r_S^1, r_S^2$ arrive at time 5. How should we schedule these requests to meet their deadlines?

\myparagraph{Straw-man Solutions}
Most inference systems default to First-Come, First-Served (FCFS) for its simplicity and fairness \cite{vllmsosp, tensorrtllm:github, 2024loongserve, agrawal2024taming, mooncake}. Under FCFS, $r_L$ executes from time 0 to 10. Requests $r_S^1, r_S^2$ wait until time 10, missing their deadlines at time 6. This demonstrates the convoy effect: short requests experience deadline violations when queued behind long-running requests. The non-preemptive nature that makes FCFS simple also makes it unsuitable for heterogeneous SLO requirements. The convoy effect is a fundamental problem in scheduling heterogeneous workloads.

The natural solution is to prioritize urgent requests. Earliest Deadline First (EDF) implements this intuition directly: always schedule the request whose deadline is soonest. Initially, EDF works as intended. At time 5, $r_S^1, r_S^2$ preempt $r_L$ since $d_S^1 = d_S^2 = 6 < d_L = 16$. However, continuous arrivals of short requests cause $r_L$ to accumulate delay. Once time exceeds $d_L$, the system enters a pathological state: $r_L$ now has a deadline $d_L <$ current time $t$, necessarily earlier than any future arrival. EDF then executes $r_L$ to completion, creating a convoy. The policy exhibits two distinct failure modes: initial starvation of long requests, followed by convoy formation for short ones. Under a constant stream of requests, both long and short requests end up missing their deadlines, and the EDF ends up performing comparable to the non-preemptive FCFS baseline (\Cref{sec:eval:sched}).

\myparagraph{Length-Aware Relative Slack (LARS)}
Slack-based scheduling naturally captures deadline urgency by tracking the time buffer before violation: $s_i = a_i + d_i - t - w_i(t)$. However, in heterogeneous workloads, raw slack values can be misleading. Two requests with identical 2-second slack face vastly different risks if one requires 5 seconds of work while another requires 100 seconds -- the longer request must survive far more scheduling decisions and potential preemptions.

LARS refines slack-based scheduling for heterogeneous workloads by scaling slack relative to work requirement: $\rho_i = s_i / w_i^{\text{total}}$. In our example, when $r_L$ arrives with absolute slack of 6 seconds but relative slack $\rho_L = 0.6$, LARS recognizes its vulnerability -- despite the seemingly comfortable buffer, it has limited slack per unit of work. Short requests arriving with $\rho_S = 1.0$ can afford to wait initially; they preempt only when their relative slack drops below the long request's.

This approach ensures all requests, regardless of length, make proportional progress toward their deadlines. Short requests still receive priority when urgent (low relative slack), but long requests aren't perpetually starved as in EDF. The result is a natural balance that avoids both convoy effects and starvation.

\subsection{Multi-Prefill Batching by Exploiting Arithmetic Intensity Slack}
\label{sec:multi-prefill-batching}

Beyond preemptive scheduling, \sysname improves system throughput by optimizing the composition of each batch. In standard chunked prefill-based scheduling \cite{agrawal2024taming}, typically only one prefill at a time. This is based on the rationale that batching multiple prefills does not improve throughput, because a single prefill chunk is already sufficient to saturate the GPU compute. However, with adaptive chunking, we observe that prefill operations at different stages of their execution have complementary resource needs -- and can benefit from batching.

\myparagraph{Arithmetic Intensity Slack in Adaptive Chunking}
We observe an opportunity to piggyback computation of short prefill requests with long prefills that allows us to compute the short prefills at a negligible cost. As \Cref{sec:adaptive-chunking}, the adaptive chunking policy dictates that we must use smaller chunks for later stages of long context prefills (when the attention cost is dominant). While the small chunk size is sufficient to saturate the attention operation, it leaves the MLP operation memory-bound. This creates \textbf{\textit{arithmetic intensity slack}}, \ie{} unused computational capacity within a batch -- which can be used to perform additional compute for negligible cost.

\myparagraph{Multi-Prefill Batching Policy}
To leverage this slack, the scheduler co-locates two prefill chunks in the same batch with complementary profiles -- packing a short, early-stage (MLP-dominant) chunk and one long, late-stage (Attention-dominant) chunk. As shown in \Cref{sec:eval:sched}, results in 1.8\myx improvement in overall system throughput.

\subsection{Dynamic Batch Packing for SLO Adherence}
\label{sec:batch-packing}

\myparagraph{Time Budget}
The batch packer constructs batches that maximize throughput while respecting a strict iteration time budget $t_{target}$. This budget, derived from decode requests' TPOT SLOs, ensures prefill operations cannot delay latency-sensitive decode tokens. Every scheduling cycle must complete within $t_{target}$ to maintain predictable decode latencies.

\myparagraph{Iterative Batch Packing}
The batch packer employs a two-phase greedy algorithm guided by a runtime performance model \Cref{alg:batch-formation}. First, it adds all active decode requests, establishing a baseline execution time $t_{decode}$. Second, it iteratively fills the remaining budget ($t_{target} - t_{decode}$) with prefill chunks in LARS priority order. For each prefill, the packer uses binary search to find the maximum chunk size that fits within the remaining budget. This process continues until the budget is exhausted or no viable chunks remain. 

Critically, this fixed-budget approach naturally implements adaptive chunking (\Cref{sec:adaptive-chunking}): early-phase prefills with empty KV caches fit large chunks within the budget, while late-phase prefills with populated KV caches are constrained to smaller chunks.

\myparagraph{Space Sharing for Multi-Prefill Batching}
As discussed in \Cref{sec:multi-prefill-batching}, the packer co-locates prefill chunks from different requests to fill the arithmetic slack and improve system throughput. Long prefills voluntarily yield a portion of their time budget based on their slack -- a request with relative slack $\rho$ uses only $(1 - \rho) \times t_{target}$, capped at a maximum yielding fraction. This mechanism creates space for short prefills without jeopardizing the long request's deadline.

Consider a long prefill with 20\% relative slack: it yields 20\% of the budget, using 16ms of the 20ms allocation. The remaining 4ms allows the packer to insert short prefills, improving overall throughput while both requests progress toward their deadlines. To prevent contention, at most one long prefill is scheduled per batch.

\myparagraph{Runtime Prediction}
The packer relies on accurate runtime predictions to make informed decisions. We use Vidur~\cite{vidur}, a performance model with <5\% prediction error. The model accounts for both chunk size and KV cache state, enabling the packer to precisely fill the time budget without violations. Through binary search over possible chunk sizes, the packer maximizes resource utilization within each scheduling cycle.

%% file: figures/parallel_strats/3dp_fig.tex
\begin{figure}[t!]
    \centering
    \includegraphics[width=0.9\linewidth]{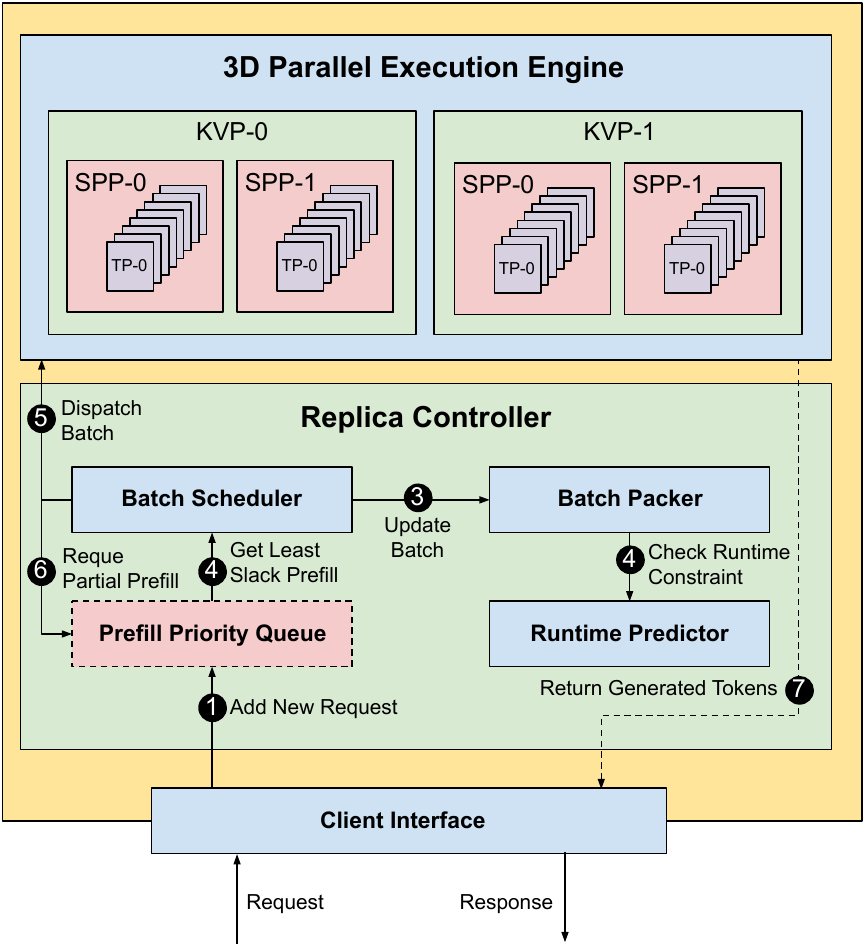}
    \caption{
    \sysname{} architecture for efficient long-context inference.
    The \emph{Replica Controller} centrally manages the request life-cycle, featuring a slack-aware \emph{Batch Scheduler} and a \emph{Batch Packer} to optimize for SLOs.
    It dispatches batches to the \emph{3D Parallel Execution Engine}, which leverages \sysname{}'s novel combination of KVP, SPP, and TP.
    }
    \label{fig:hld}    
\end{figure}

%% file: 5-optimizations.tex
\section{Implementation}
\label{sec:optimizations}

\sysname{} extends the Sarathi-Serve framework \cite{agrawal2024taming} to tackle multi-million token context requests.
Unlike vLLM and Sarathi-Serve, which incur overhead from centralized schedulers as sequence length grows, we reduce communication by replicating sequence state across the scheduler and GPU workers.

We replace Ray~\cite{ray} with ZeroMQ~\cite{zmq} for scheduler-worker communication, eliminating GIL contention as we scale to hundreds of workers.
We also integrate FlashInfer~\cite{flashinfer} kernels to distribute work across both query and KV tokens, optimizing chunked prefill for long contexts.
To meet strict latency targets with small prefill chunks, we implement the model execution engine’s critical path in C++ using PyBind, ensuring seamless integration with the Python codebase.

%% file: 6-eval.tex
\section{Evaluation}
\label{sec:eval}

\vspace{-0.1em}

\subsection{Evaluation Setup}

\input{figures/experiments/e2e/a100/a100_e2e_ttft_fig}
\input{figures/experiments/e2e/a100/a100_e2e_tbt_fig}

\myparagraph{Baselines}
We compare our system against the \sota{} long-context LLM inference serving systems, \ls{} \cite{2024loongserve} and vLLM \cite{vllmsosp}.
Note that, for context lengths greater than 32K, vLLM defaults to the  Sarathi-Serve scheduler \cite{agrawal2024taming}.
Thus, we refer to this baseline as Sarathi.
We consider two chunk sizes for the Sarathi scheduler: 512 and 2048. We also consider DistServe \cite{distserve2024} and SplitWise \cite{patel2023splitwise}, however, these systems run out of memory due to activation memory bottleneck as discussed in \Cref{sec:limits-conventional}. To our knowledge, there are no publicly available systems that directly tackle convoy effects in long context inference. Finally, we evaluate \sysname{} variant that replaces the LARS request prioritization and multi-prefill batching with standard FCFS/EDF/LRS scheduling while retaining all other proposed mechanisms.

\myparagraph{Models and datasets}
We use \llamaS{} and \llamaL{} with RoPE~\cite{su2024roformer} scaling to support up to 10M tokens. 
Currently, there are no publicly available long-context LLM datasets available that span millions of tokens.
Previous systems use L-Eval~\cite{leval} and LV-Eval~\cite{lveval} for long context evaluations. These datasets were created to evaluate long-context abilities of LLMs and predominantly contain short form question, with extremely small decode lengths. For instance, the median output length in L-Eval is 47 tokens as opposed to 415 in ShareGPT4 ~\cite{wang2023openchat} --- which is based on actual real-world user interactions with GPT4.

To perform more realistic evaluations, we construct the \textbf{\emph{\sysname{}-SWE}} trace using the Gemini-Flash-1.5B model~\cite{team2024gemini}, inspired by LLM-enabled software engineering tools that have recently gained popularity.
We focus on two common engineering tasks: code review for pull requests and GitHub issue resolution.
From the top 1,000 most-starred GitHub repositories with permissive licenses (Apache or MIT), we select those with token counts between 100K and 1M.
We extract the 100 most recent issues and merged PRs per repo and prompt Gemini to solve them referencing the codebase.

This yields interactions with prefill lengths of 393K (P50) and 839K (P90) tokens and decode lengths of 518 (P50) and 808 (P90). 
To simulate a realistic request mix, we combine these long-context examples with the ShareGPT4 trace~\cite{wang2023openchat}, which consists of real GPT-4 conversations capped at 8K tokens.
We test \sysname{} under various ratios of long and short-context requests.

\myparagraph{Hardware}
We evaluate \sysname across two hardware setups.
For the \llamaS{} model, we use a setup with two DGX-A100 servers~\cite{a100azure}.
While for \llamaL{}, we use a 128-GPU cluster with 16 DGX-H100 servers~\cite{h100azure}.
In both setups, each server has 8 GPUs with 80GB of high bandwidth memory.
The GPUs within a server are connected with NVLINK.
Cross-server connection is via InfiniBand.

\input{figures/experiments/e2e/h100/h100_e2e_ttft_fig}

\subsection{Capacity Evaluation}
\label{sec:eval:cap}
We begin by evaluating how \sysname{} performs under varying loads compared to existing approaches for \llamaS{} model on the A100 cluster.
Our capacity evaluation focuses on two key metrics: TTFT and TPOT, as these directly impact user experience in interactive scenarios.

To evaluate capacity systematically, we designed two workload scenarios:
(1) a baseline with only short-context requests (\ie{}, ShareGPT4) and
(2) a mixed workload containing 5\% long-context requests (128K--1M tokens).
We vary the system load from 0.25 to 1.75 queries per second (QPS) and compare \sysname{} against \ls{} (TP-2, CP-4) and Sarathi (TP-8, PP-2).
For fairness, we configure \sysname{} with similar configuration (TP-8, SPP-2).

\myparagraph{Baseline Performance}
In the scenario with only short requests (\Cref{fig:e2e:a100:ttft:short}), all systems exhibit comparable performance at low loads (0.25 QPS). 
However, as load increases, \ls{}'s performance degrades considerably, which we attribute to resource fragmentation.
At 1.75 QPS, \ls{}'s P90 TTFT increases dramatically, while \sysname{} maintains consistent latency. Furthermore, \sysname achieves considerably better latency compared to Sarathi due to \sysname's SPP, which helps reduce TTFT.

\myparagraph{Long Query Performance}
\Cref{fig:e2e:a100:ttft:long} shows significant benefits for \sysname{} with long-context requests.
At 0.75 QPS, \sysname{} achieves a 30\myx median TTFT improvement over \ls{}.
Sarathi and \sysname{}-FCFS quickly degrade due to the convoy effect.
Even at 1.25 QPS, \sysname{} maintains acceptable TTFT latencies, offering ~5\myx{} higher effective capacity than the baselines.
Some baseline systems fail to complete requests within the 60-minute profiling window due to convoy effect, resulting in truncated CDFs.

\myparagraph{Decode Performance}
\Cref{fig:e2e:a100:tpot} shows that \ls{} experiences ~5\myx{} higher TPOT latencies than \sysname{}, even at high loads without long requests, due to resource fragmentation.
With long requests, \sysname{} achieves comparable or better TPOT while processing significantly more requests with an order of magnitude lower TTFT.
Even Sarathi, optimized for low decode latency, reaches TPOTs as high as 1 second due to its static chunking approach, which increases costs for processing later chunks in long sequences.
In contrast, \sysname{}'s adaptive chunking maintains consistent performance across varying sequence lengths.

\subsection{3D Parallel Performance}

With \sysname{}'s baseline established, we evaluate 3D parallelism that combines tensor, stream pipeline and KV parallelism. For this experiment we use \llamaL{} on a H100 cluster.
We compare two setups with equal resource budgets:
(1) a 2D configuration (SPP-8) and
(2) a 3D configuration (SPP-4, KVP-2), both using TP8.
We run a mixed workload, including 5\% long-context (2M token) requests, scaled from the \sysname{}-SWE trace.

\Cref{fig:e2e:h100:ttft:long} shows TTFT distributions under varying loads.
At lower request rates (0.25 and 0.75 QPS), both configurations perform similarly, with nearly identical CDF curves.
At higher loads (1.25 and 1.75 QPS), a trade-off emerges:
the 3D parallel setup offers slightly lower peak throughput due to the higher SPP degree in the 2D case, which is more communication-efficient than KVP and better accelerates prefill.
Despite this, both configurations maintain similar median latencies.

\Cref{fig:e2e:h100:tpot:long} shows the strength of 3D parallel in the decode phase.
At high load (1.75 QPS), the 3D setup reduces TPOT by over 2\myx{} at both P50 and P90.
Even small prefill chunks can delay co-batched decode requests, especially with 2M-token sequences and large models.
KVP mitigates this by distributing KV cache reads, reducing decode latency.

This confirms a core design goal of \sysname{}’s 3D parallelism:
balancing prefill throughput with decode responsiveness.
While the 2D setup favors prefill speed, 3D parallelism delivers more consistent end-to-end latency—critical for real-world deployments.
It retains the benefits of SPP while combining the strengths of both approaches.

\input{figures/experiments/sched_abl/scheduler_abl_figure}

\subsection{Effectiveness of \sysname{} Scheduler}
\label{sec:eval:sched}

We isolate the performance gains from \sysname{}'s scheduling policies by comparing it to traditional scheduling policies.
\Cref{fig:abl:sched} shows the TTFT distributions for four approaches:
FCFS, EDF, LARS (without multi-prefill batching), and \sysname{}'s scheduler with all optimizations enabled.
The evaluation uses \llamaS{} on A100 GPUs in TP8-SPP2 configuration with a mixed workload of 5\% long-context requests.

At low load (0.25 QPS), all policies show similar median latency but differ in tail behavior.
However, at high load (1.75 QPS), the differences become more pronounced.
FCFS performs poorly due to unmitigated convoy effect from long requests.
Despite its success in latency-sensitive systems, EDF struggles here.
While effective at low loads, EDF's performance degrades at higher loads, resembling FCFS behavior.
This occurs because EDF defers long requests until their deadlines become unfeasible, causing them gain highest priority once they pass their deadlines as discussed in \Cref{sec:prioritization}.

We also compare \sysname{} with multi-prefill batching to vanilla LARS. While both of these setups significantly outperform the FCFS and EDF baselines by mitigating convoy effect, we up to 1.8\myx lower median latency with \sysname{} compared to the vanialla LARS setup due to more effective GPU utilization enabled by multi-prefill batching.

\myparagraph{Sensitivity to Long Request Mix}
Figure~\ref{fig:banner:bars} shows how TTFT degrades as the fraction of long requests increases from 0\% to 5\%. 
The baseline system exhibits superlinear degradation --- even with 1\% long requests \ls shows 8\myx higher latency, while at 5\% it exhibits 30\myx P50 and 174\myx P90 higher latency due to convoy effect.
In contrast, \sysname gracefully maintains the P90 latency under 10 seconds even with 5\% long request mix.

\subsection{Alternate Scheduling Approaches: Multiuple Request Pools}
A common industry technique to tackle with heterogeneity when serving models with moderate context lengths (64-128K) is to create separate pools for short and long requests.
While \ls{} dynamically creates similar pools based on prefill lengths, it does not guarantee the availability of dedicated resources for all short requests.
To evaluate the effectiveness of this approach, we implement a version of \ls{} with a \emph{reserved pool} specifically for short request processing, as shown in \Cref{fig:experiments:longservepp}.

We compare \sysname{} to this baseline using the same setup as \Cref{sec:eval:cap}, reserving two of eight CP instances for short requests (<8192 tokens) and the rest for long requests.
Each pool uses the standard \ls{} scheduler.
This reservation increases contention for long prefills, leading to up to 20\% lower completions for long requests  compared to \sysname{}, and 10\% lower than default \ls{}. For the decodes, \ls with reservation achieves slightly lower TPOT compared to \ls as an artifact of overall lower ingestion (prefill) rate. \sysname consistently achieve lower decode latency compared to be both the variants of \ls. Thus, creating separate pools for requests of different length does not solve the fundamental problem of convoy effect while hurting throughput due to fragmentaion.

\input{figures/experiments/longservepp/longservepp_comparison_fig}

%% file: figures/experiments/e2e/a100/a100_e2e_ttft_fig.tex
\begin{figure*}[tbh]
    \centering
    \begin{subfigure}[b]{\linewidth}        
        \centering
        \includegraphics[width=0.87\linewidth]{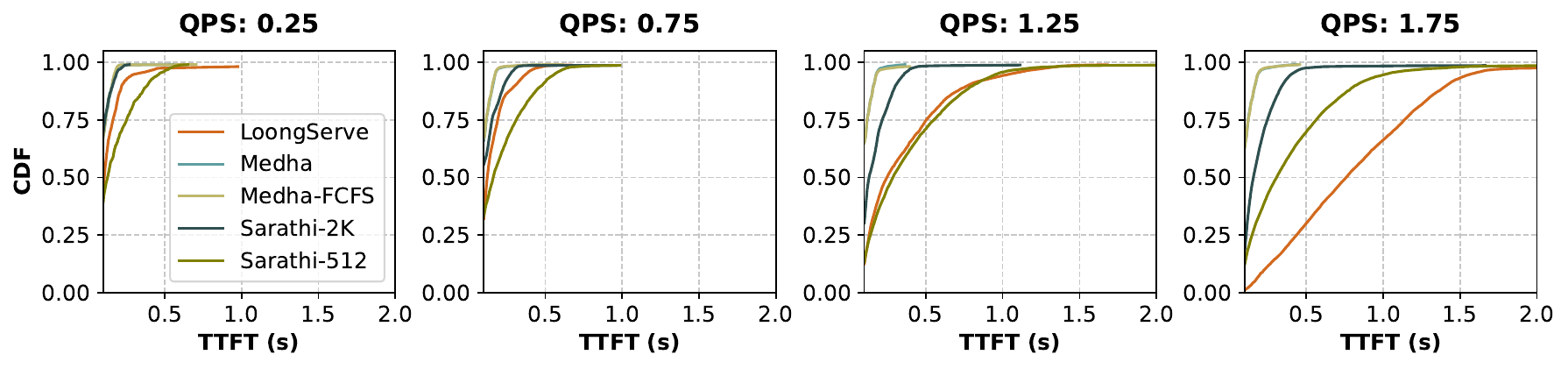}
        \caption{For short-context workloads from ShareGPT4, \sysname{} maintains consistently low latency even at high QPS.}
        \label{fig:e2e:a100:ttft:short}
    \end{subfigure}
    \begin{subfigure}[b]{\linewidth}
        \centering
        \includegraphics[width=0.87\linewidth]{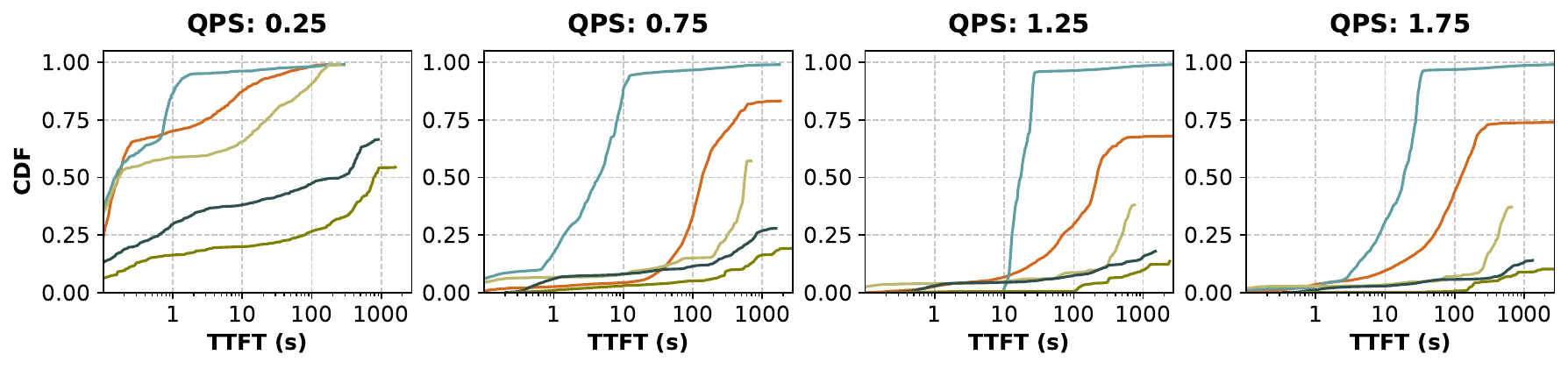}
        \caption{For ShareGPT4 with 5\% long requests, \sysname{} achieves up to 30\myx{} lower median TTFT, demonstrating effective mitigation of HOL blocking}
        \label{fig:e2e:a100:ttft:long}
    \end{subfigure}
    \caption{TTFT latency distribution under varying load conditions for \llamaS{} on two servers with a total of 16 A100 GPUs.}
    \label{fig:e2e:a100:ttft}
\end{figure*}

%% file: figures/experiments/e2e/a100/a100_e2e_tbt_fig.tex
\begin{figure}[t!]
    \centering
    \begin{subfigure}[b]{\linewidth}        
        \centering
        \includegraphics[width=0.9\textwidth]{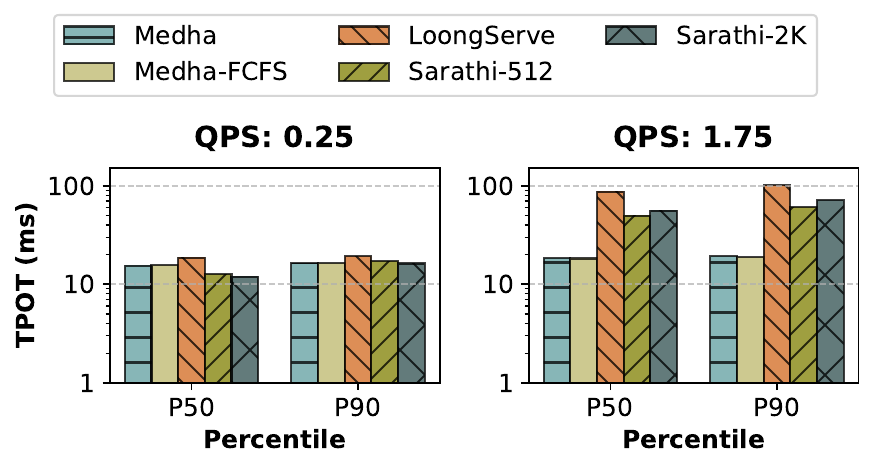}
        \caption{ShareGPT4.}
        \label{fig:e2e:a100:tpot:short}
    \end{subfigure}
    \begin{subfigure}[b]{\linewidth}
        \centering
        \includegraphics[width=0.9\textwidth]{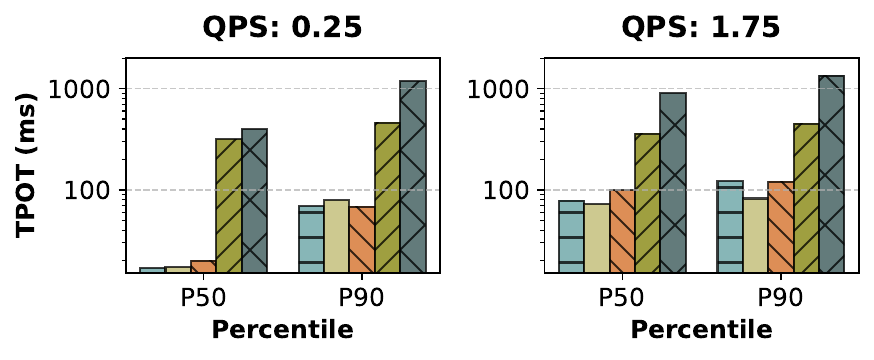}
        \caption{ShareGPT4 with 5\% long requests.}
        \label{fig:e2e:a100:tpot:long}
    \end{subfigure}
    \caption{Decode latency for \llamaS{} on 16 A100s.
    Due to adaptive chunking, \sysname{} maintains low decode latency while other chunked prefill-based systems suffer from high latency.}
    \label{fig:e2e:a100:tpot}
\end{figure}

%% file: figures/experiments/e2e/h100/h100_e2e_ttft_fig.tex
\begin{figure*}[t!]
    \centering
    \includegraphics[width=0.9\linewidth]{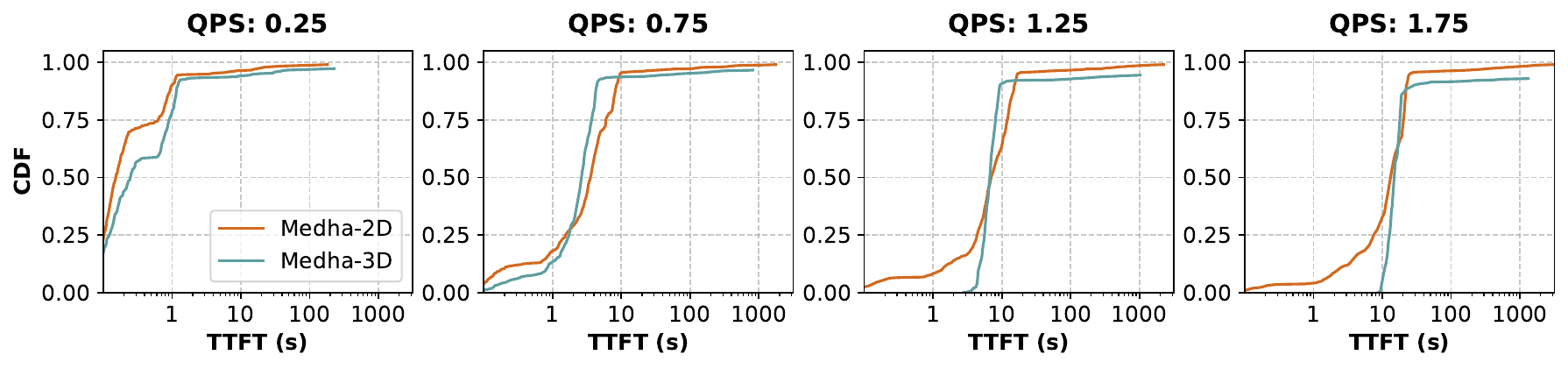}
    \caption{Prefill performance comparison of parallelization strategies for \llamaL{} on 8 64 H100 GPUs running ShareGPT4 with 5\% long requests.}
    \label{fig:e2e:h100:ttft:long}
\end{figure*}

\begin{figure}[t!]
        \centering
        \includegraphics[width=0.9\linewidth]{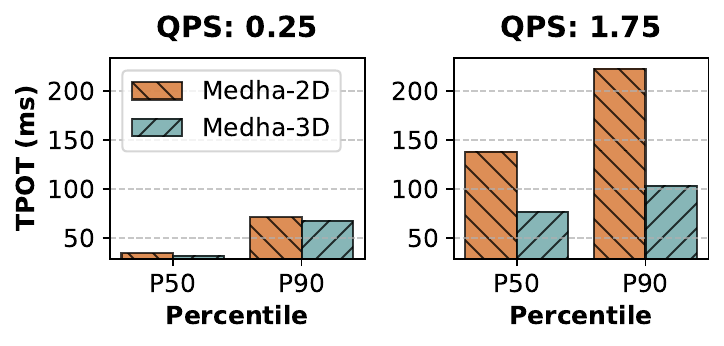}
    \caption{\sysname{}-3D (SPP+TP+KVP) maintain comparable TTFT performance \sysname-2D (SPP+TP) and but enable 2\myx better decode performance by distributing KV cache reads and reducing prefill-decode interference.}
    \label{fig:e2e:h100:tpot:long}
\end{figure}

%% file: figures/experiments/sched_abl/scheduler_abl_figure.tex
\begin{figure}[t!]    
    \centering
    \includegraphics[width=0.9\linewidth]{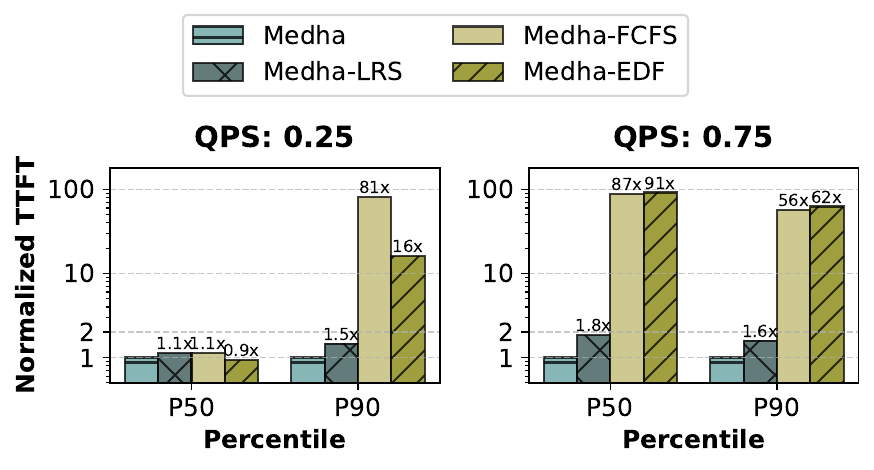}
    \caption{Impact of different scheduling policies on normalized TTFT latency. Even compared to our modified LRS policy, \sysname scheduler achieves (1.6--1.8\myx) lower latency, demonstrating the effectiveness of \sysname's prefill-prefill batching technique.
}
    \label{fig:abl:sched}
\end{figure}

%% file: figures/experiments/longservepp/longservepp_comparison_fig.tex
\begin{figure}[t!]
    \centering
    \begin{subfigure}[b]{0.49\linewidth} 
    \includegraphics[width=0.95\linewidth]{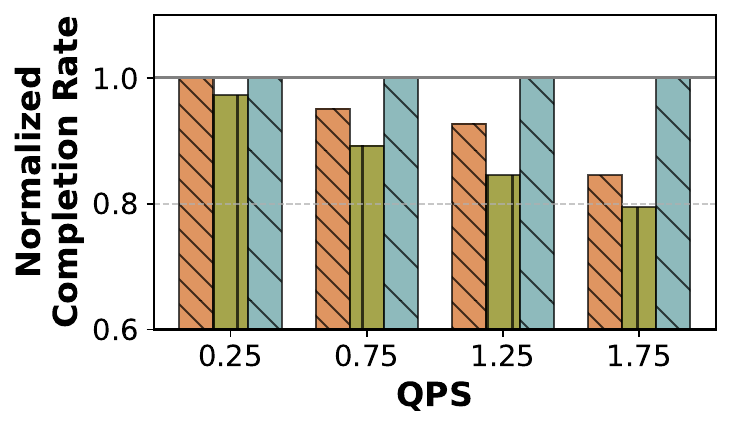}
    \caption{Normalized completion rate.}
    \end{subfigure}
    \begin{subfigure}[b]{0.49\linewidth} 
    \includegraphics[width=\linewidth]{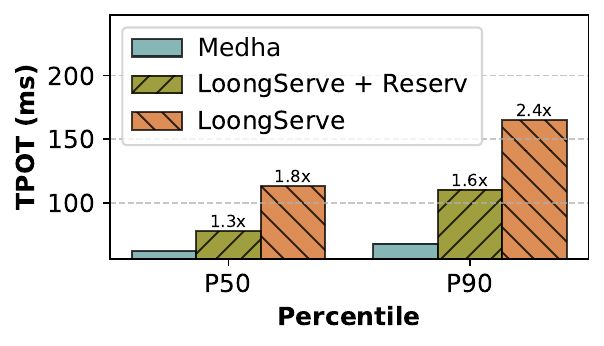}
    \caption{Decode latency at 0.75 QPS.}
    \end{subfigure}
    \caption{
        Impact of pool fragmentation on long requests for \llamaS{} on 16-A100 with 5\% long requests.
        \sysname{} maintains maximum throughput and lowest latency.
        Adding a dedicated reserved pool to \ls{} (+Reserv) to mitigate HOL blocking for short requests fragments resources and further degrades overall completion rate for long requests compared to both standard \ls{}.
    }
    \label{fig:experiments:longservepp}    
\end{figure}

%% file: 8-related.tex
\section{Related Work}
\label{section:relatedwork}

\myparagraph{LLMs for long context}
Recent research has focused on effectively training and serving long-context LLM models. Some propose new attention parallelism techniques as more efficient solutions to enable long context~\cite{2021sequence, liu2023ring,brandon2023striped}.
We discuss and compare them in detail in \Cref{sec:background-condensed,sec:eval}.
A similar idea to SPP, called token-parallelism, was used in TeraPipe~\cite{terapipe} to parallelize the different micro-batches of a mini-batch along the token dimension in order to reduce pipeline bubbles and improve throughput during training. Recently, Mooncake \cite{mooncake} --Kimi.ai's proprietary serving system, a work parallel to ours, concurrently proposed use of this technique to reduce TTFT latency during inference. Note that, while Mooncake explores use of chunked prefills to accelerate long prefill computation, it does not address convoy effect. To the best of our knowledge, \sysname is the first system to leverage chunked prefills for preemptive scheduling to tackle heterogeneity in long context serving.

\myparagraph{Request scheduling}
Efficient request scheduling has been extensively studied~\cite{fu2024efficient,sheng2023fairness,sun2024llumnix,qiu2024muserve,liu2024andes,aiops2024qiu,wu2023fast}, but existing approaches have notable limitations when addressing long-context requests.
For example, SRTF scheduling~\cite{fu2024efficient,aiops2024qiu} reduces median latency but leads to starvation of long requests due to lack of preemption.
LoongServe\cite{2024loongserve} supports space sharing among concurrent long requests but lacks preemption and time-sharing, resulting in significant HOL delays, especially under FCFS scheduling.
Fairness-focused schedulers like \cite{sheng2023fairness} emphasize equitable resource distribution among clients but fail to address strict latency SLOs.
In contrast, \sysname{} introduces a slack-based fine time sharing scheduling policy with prefill-prefill batching, enabling efficient mixing of long and short requests to meet latency SLOs.%

%% file: 9-conc.tex
\section{Conclusion}
\label{sec:conclusion}

This work demonstrates that the convoy effect, long understood in operating systems, is a critical but overlooked challenge in long-context LLM serving. Traditional non-preemptive systems fail to tackle the extreme heterogeneity caused by the quadratic attention cost, as a result a single long request can drastically degrade service for hundreds of short queries. Our results show that, that with careful co-design of parallelism strategies and scheduling policies preemption can be both practical and effective. As context windows of state-of-the-art LLMs continues to grow, the heterogeneity problem will only intensify, making preemptive scheduling a requirement rather than an optimization. \\sysname{} shows that with careful system design, we can effectively serve long-context LLM workloads at scale.

%% file: 10-appendix.tex
\newpage
\section{Scheduling Algorithms}
\label{appendix:algorithms}

This appendix presents the detailed algorithms used in \sysname's scheduling and batch packing mechanisms. We provide pseudocode for the adaptive batching process, chunk size selection, and prefill time estimation procedures.

\subsection{Batch Formation and Packing}

The batch formation process operates in two phases: first packing latency-sensitive decode requests, then filling remaining capacity with adaptively-chunked prefill requests. Algorithm~\ref{alg:batch-formation} presents the core batching logic.

\begin{algorithm}[htbp]
\caption{SLO-Aware Adaptive Batching}
\label{alg:batch-formation}
\begin{algorithmic}[1]
\Require $\mathcal{Q}$: Queue of pending requests (prefill and decode)
\Require $t_{target}$: Target batch execution time (e.g., 20ms from TPOT SLO)
\Ensure $\mathcal{B}$: Batch of (request, num\_tokens) pairs
\Procedure{FormNextBatch}{$\mathcal{Q}, t_{target}$}
    \State $\mathcal{B} \gets \emptyset$ \Comment{Batch of (request, tokens) pairs}
    \State $t_{pred} \gets 0$ \Comment{Predicted batch execution time}
    
    \State \textbf{Phase 1: Pack decode requests}
    \State $\mathcal{D} \gets$ GetDecodeRequests($\mathcal{Q}$)
    \State $\mathcal{B} \gets \mathcal{B} \cup \mathcal{D}$
    \State $t_{pred} \gets$ PredictTime($\mathcal{B}$)
    
    \State \textbf{Phase 2: Fill with prefill chunks}
    \State $\mathcal{P} \gets$ GetPrioritizedPrefills($\mathcal{Q}$) \Comment{LARS-ordered}
    \While{$\mathcal{P} \neq \emptyset$ and $t_{pred} < t_{target}$}
        \State $r \gets$ Pop($\mathcal{P}$)
        \State $n \gets$ GetChunkSize($r, t_{pred}, t_{target}, \mathcal{B}$)
        \If{$n > 0$}
            \State $\mathcal{B} \gets \mathcal{B} \cup \{(r, n)\}$
            \State $t_{pred} \gets$ PredictTime($\mathcal{B}$)
        \Else
            \State Requeue($r, \mathcal{P}$)
        \EndIf
    \EndWhile
    \State \Return $\mathcal{B}$
\EndProcedure
\end{algorithmic}
\end{algorithm}

The algorithm prioritizes decode requests to meet their strict TPOT requirements, then iteratively adds prefill chunks based on LARS priority ordering. The target batch time $t_{target}$ serves as a hard constraint derived from decode SLOs.

\subsection{Adaptive Chunk Size Selection}

The chunk size selection mechanism implements both the adaptive chunking policy and space-sharing optimization. Algorithm~\ref{alg:chunk-size} shows how chunk sizes are determined based on current batch composition and request urgency.

\begin{algorithm}[h]
\caption{Adaptive Chunk Size Selection}
\label{alg:chunk-size}
\begin{algorithmic}[1]
\Require $r$: Prefill request with KV cache size $r.kv\_size$
\Require $t_{current}$: Current predicted batch execution time
\Require $t_{target}$: Target batch execution time limit
\Require $\mathcal{B}$: Current batch composition
\Require $\rho_{max}$: Maximum space-sharing fraction (e.g., 0.4)
\Ensure Chunk size in tokens, or 0 if request cannot be scheduled
\Procedure{GetChunkSize}{$r, t_{current}, t_{target}, \mathcal{B}$}
    \State \textbf{Safety constraint:} Prevent multiple long prefills
    \If{IsLong($r$) $\land$ ContainsLongPrefill($\mathcal{B}$)}
        \State \Return 0
    \EndIf
    
    \State \textbf{Space sharing:} Long requests yield time for urgency
    \State $\rho \gets$ GetRelativeSlack($r$) \Comment{$\rho = s(t) / w^{\text{total}}$}
    \State $\rho \gets \min(\rho_{max}, \max(0, \rho))$ \Comment{Cap at max sharing}
    
    \State \textbf{Calculate effective budget:}
    \State $t_{effective} \gets t_{target} \cdot (1 - \rho)$
    \State $t_{budget} \gets t_{effective} - t_{current}$
    
    \If{$t_{budget} \leq 0$}
        \State \Return 0
    \EndIf
    
    \State \textbf{Binary search for maximum chunk:}
    \State \Return BinarySearchChunk($r, t_{budget}$)
\EndProcedure
\end{algorithmic}
\end{algorithm}

The algorithm implements two key policies: (1) space-sharing where requests with high relative slack $\rho$ yield time to more urgent requests, and (2) finding the maximum chunk size that fits within the allocated time budget.

\subsection{Prefill Time Estimation}

Accurate estimation of remaining prefill time is crucial for LARS scheduling. We precompute a cache of prefill times for various sequence lengths, accounting for the adaptive chunking policy. Algorithm~\ref{alg:prefill-time} presents both the offline precomputation and runtime estimation procedures.

\begin{algorithm}[htbp]
\caption{Remaining Prefill Time Estimation}
\label{alg:prefill-time}
\begin{algorithmic}[1]
\State \textbf{Offline Precomputation:}
\Require $L_{max}$: Maximum sequence length to cache (e.g., 1M tokens)
\Require $\Delta$: Granularity of cache entries (e.g., 1K tokens)
\Ensure $\mathcal{C}$: Cache mapping sequence length to total prefill time
\Procedure{BuildPrefillCache}{$L_{max}, \Delta$}
    \State $\mathcal{C} \gets \{0 \mapsto 0\}$ \Comment{Cache: tokens $\rightarrow$ time}
    \For{$\ell = \Delta$ to $L_{max}$ step $\Delta$}
        \State $\mathcal{C}[\ell] \gets$ SimulatePrefill($\ell$)
    \EndFor
    \State \Return $\mathcal{C}$
\EndProcedure

\State
\Require $L$: Total sequence length to simulate
\Ensure Total time to prefill $L$ tokens with adaptive chunking
\Procedure{SimulatePrefill}{$L$}
    \State $t_{total} \gets 0$
    \State $\ell_{processed} \gets 0$
    
    \While{$\ell_{processed} < L$}
        \State $kv_{size} \gets \ell_{processed}$ \Comment{Current KV cache size}
        \State $(c, t_c) \gets$ GetOptimalChunk($kv_{size}, t_{target}$)
        \If{$c = 0$}
            \State \textbf{break}
        \EndIf
        \State $t_{total} \gets t_{total} + t_c$
        \State $\ell_{processed} \gets \ell_{processed} + c$
    \EndWhile
    \State \Return $t_{total}$
\EndProcedure

\State
\State \textbf{Runtime Estimation:}
\Require $\ell_{total}$: Total tokens in the request
\Require $\ell_{processed}$: Tokens already processed
\Ensure Estimated time to complete remaining tokens
\Procedure{GetRemainingTime}{$\ell_{total}, \ell_{processed}$}
    \State $t_{total} \gets$ LookupCache($\mathcal{C}, \ell_{total}$)
    \State $t_{done} \gets$ LookupCache($\mathcal{C}, \ell_{processed}$)
    \State \Return $t_{total} - t_{done}$
\EndProcedure
\end{algorithmic}
\end{algorithm}

The simulation accounts for how chunk sizes decrease as the KV cache grows, reflecting the shift from MLP-dominant to attention-dominant computation. The \textsc{SimulatePrefill} procedure uses the Vidur simulator~\cite{vidur} to model execution times accurately.

\input{figures/experiments/mbu_mfu_scaling_fig_combined}

\section{Scaling Properties of Parallelism Strategies}
\label{appendix:scaling}

This appendix provides detailed scaling analysis of \sysname's parallelism strategies: Stream Pipeline Parallelism (SPP) for prefill acceleration and KV-Cache Parallelism (KVP) for decode latency bounding.

\subsection{Stream Pipeline Parallelism Scaling}

\input{figures/experiments/pp_scaling/pp_scaling_ttft_combined}

\input{figures/experiments/pp_scaling/pp_scaling_tbt_combined}

\Cref{fig:sppscaling:ttft} demonstrates the scaling efficiency of Stream Pipeline Parallelism across different model sizes and sequence lengths. We evaluate \sysname 2D (SPP+TP) configurations against baseline approaches for long-context prefill processing.

For Llama-3 8B (~\Cref{fig:sppscaling:ttft:sevenb}), \sysname achieves near-linear scaling up to 128 H100 GPUs. The efficiency remains above 70\% even at the highest parallelism degrees, demonstrating effective overlap of computation and communication. Notably, \sysname outperforms ring attention approaches by 60\% due to eliminating the sequential dependency bottleneck.

Scaling to Llama-3 70B (~\Cref{fig:sppscaling:ttft:seventyb}) shows even stronger benefits. The larger model's increased compute density better amortizes pipeline startup costs, achieving 85\% scaling efficiency at SPP degree 8.

\Cref{fig:sppscaling:tbt} examines the decode latency implications of SPP scaling. Due to it's communication efficient nature, SPP only marginally affects decode performance due to pipeline depth. With SPP degree 8, decode latency increases by only 16\%.

\subsection{KV-Cache Parallelism Impact on Decode Performance}

Figure~\ref{fig:kvpscaling:decode} shows KVP's effectiveness in bounding decode latency. For 10M-token contexts on Llama-3 8B, KVP with degree 4 reduces TPOT by 40\% in decode-only batches. The scaling is sub-linear due to communication overhead, but the latency reduction is crucial for meeting decode SLOs with long contexts.

\subsection{TTFT-TPOT Trade-off Analysis}

\input{figures/experiments/cp_prefill_scaling/cp_scaling_prefill_tfft_all_fig}

We sweep the space of various chunk sizes for the chunked prefill, and also vary $p_{kvp}$, while keeping $p_{spp}=4$.
\Cref{fig:kvpscaling:prefill:ttft} shows the results on \llamaS.
For a given $p_{kvp}$, increasing the chunk size, reduces TTFT (prefill latency), since it requires fewer iterations.
At the same time, it increases TBT, since each batched iteration takes longer to execute.
Therefore, for sequence length 1M with $p_{kvp}=1$, the green line shows the left-most triangle at largest chunk size, and the right-most triangle at the smallest chunk size. 
For a given chunk size, increasing $p_{kvp}$ helps reduce both TTFT and TBT in most cases, thus helping reach more optimal points in this trade-off space.
Indeed, lower $p_{kvp}$ achieves better TTFT latency in cases with lower arithmetic intensity (due to small chunk size), as exemplified by the right-most points for 1M context length. As we increase the arithmetic intensity (\eg{}, 2M context length), we see increasing $p_{kvp}$ achieving the same performance for the smallest chunk size, and, finally, decreasing TTFT for 4M context length.

\subsection{Resource Utilization Efficiency}

A key measure of \sysname{}'s effectiveness is its ability to maintain high throughput while scaling to large parallelism degrees.
We evaluate this using hardware utilization metrics Model FLOPS Utilization (MFU) and Model Bandwidth Utilization (MBU)~\cite{chowdhery2023palm, llmperfblog}.
In LLM inference, prefill phases are compute-bound while decode phases are memory-bound~\cite{patel2023polcapoweroversubscriptionllm,patel2023splitwise}.
\Cref{fig:utilization:sppmfu} shows the MFU for \sysname{} in the prefill phase (2D SPP+TP), while \Cref{fig:utilization:kvpmbt} shows the MBU for the decode phase (2D KVP+TP).
For \llamaL{}, we achieve 50--60\% MFU across configurations, improving for longer sequences.
Even at the scale of 128 GPUs, we achieve over 50\% MFU.
Examining MBU, \Cref{fig:utilization:kvpmbt} shows that \sysname{}'s KVP implementation achieves up to 92\% MBU in optimal configurations, allowing consistent decode performance even with extremely long contexts.

%% file: figures/experiments/mbu_mfu_scaling_fig_combined.tex
\begin{figure*}[ht]
    \begin{minipage}[t]{0.58\textwidth}
        \centering
        \includegraphics[width=\linewidth]{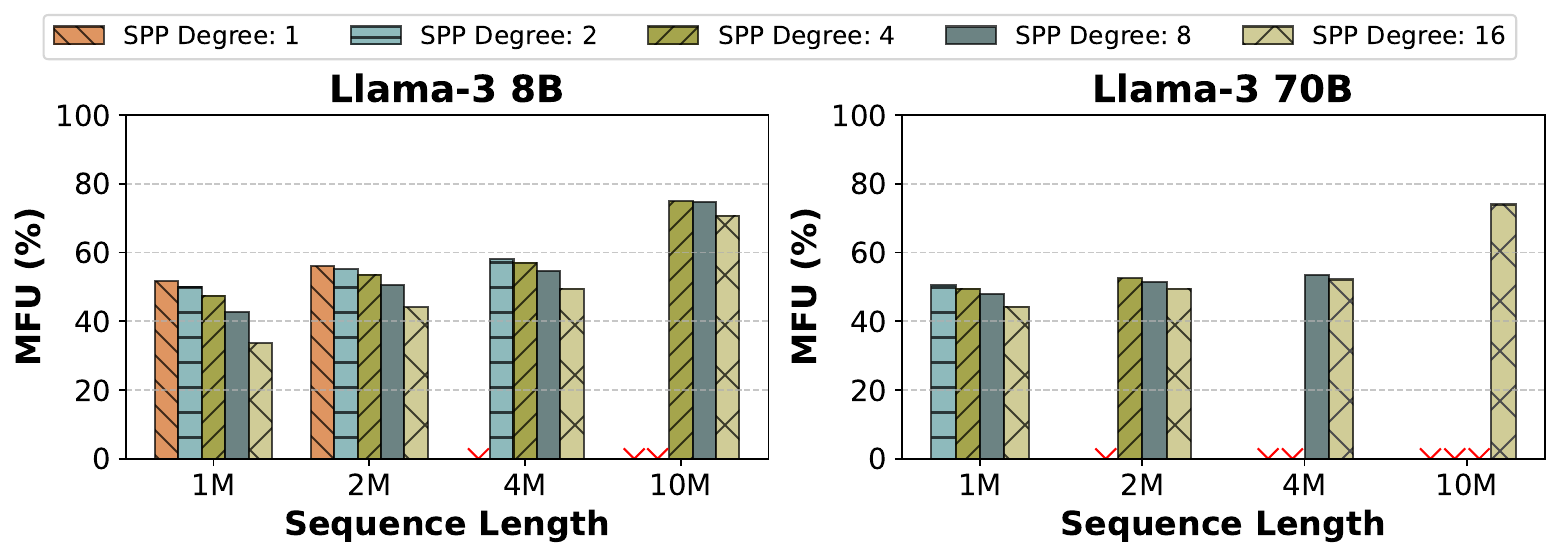}    
        \caption{
            Model FLOPS Utilization~\cite{chowdhery2023palm} (MFU) for \sysname{} 2D (TP+SPP).
            It achieves 50-60\% utilization across sequence lengths and parallelism degrees.
        }
        \label{fig:utilization:sppmfu}
    \end{minipage}
    \hspace{0.5em}
    \begin{minipage}[t]{0.39\textwidth}
        \centering
        \includegraphics[width=\linewidth]{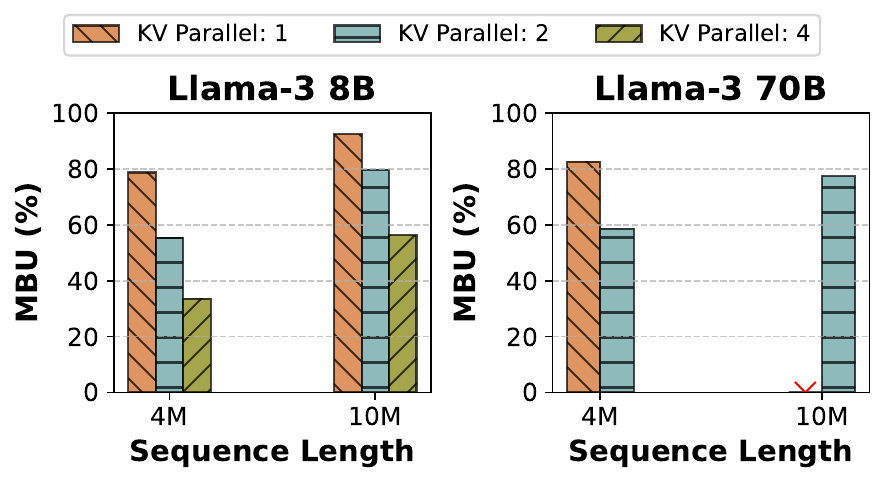}    
        \caption{
            Model Bandwidth Utilization (MBU) for \sysname{} 2D (TP+KVP).
        }
        \label{fig:utilization:kvpmbt}
    \end{minipage}
\end{figure*}

%% file: figures/experiments/pp_scaling/pp_scaling_ttft_combined.tex
\begin{figure*}[tbh]
    \centering
    \begin{subfigure}[b]{0.49\linewidth}        
        \centering
        \includegraphics[width=\textwidth]{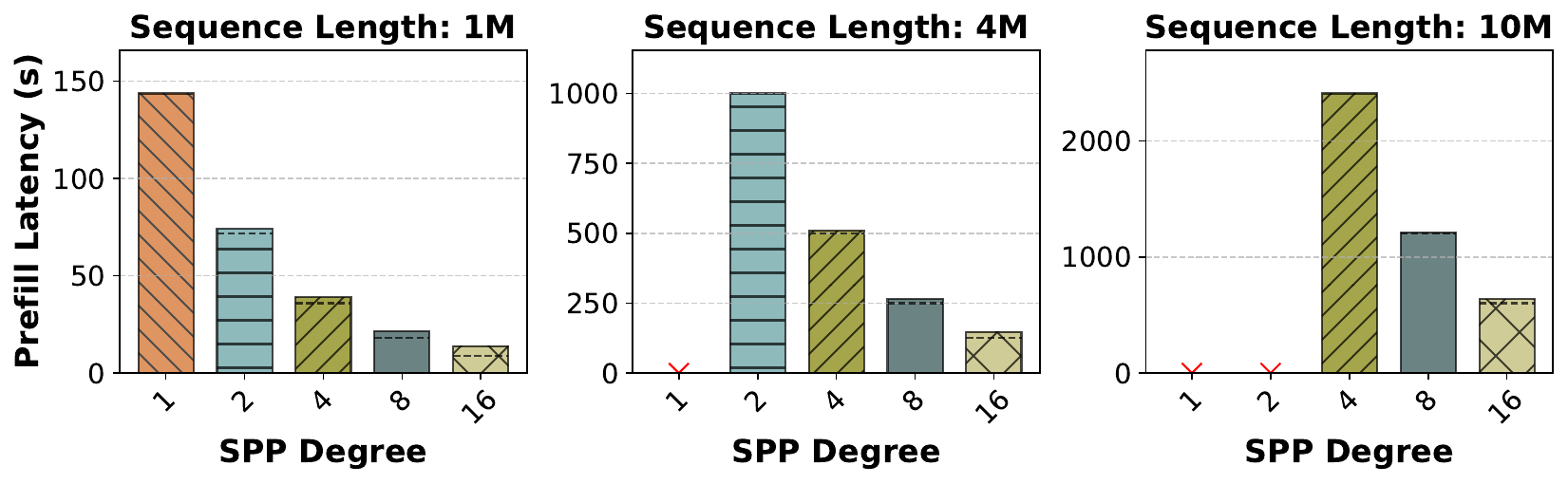}
        \caption{\llamaS.}
        \label{fig:sppscaling:ttft:sevenb}
    \end{subfigure}
    \begin{subfigure}[b]{0.49\linewidth}
        \centering
        \includegraphics[width=\textwidth]{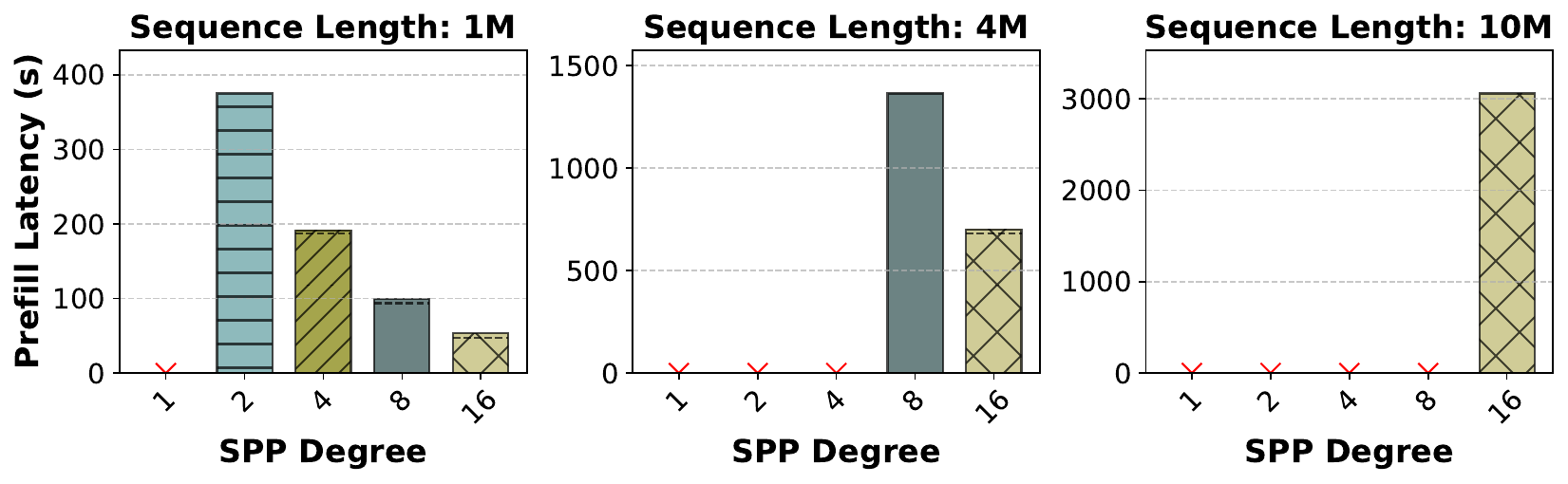}
        \caption{\llamaL.}
        \label{fig:sppscaling:ttft:seventyb}
    \end{subfigure}
    \caption{
    Scaling efficiency of \sysname{} 2D (SPP+TP) for long-context prefill processing.
    \sysname{} 2D reduces TTFT near-linearly (80\%+ scaling efficiency) as the SPP degree increases to operate with up to 128 H100 GPUs.
    Red crosses are infeasible settings due to memory limitations.}
    \label{fig:sppscaling:ttft}
\end{figure*}

%% file: figures/experiments/pp_scaling/pp_scaling_tbt_combined.tex
\begin{figure*}[ht]
    \begin{minipage}[t]{0.39\textwidth}
    \begin{subfigure}[b]{0.45\linewidth}      
        \centering
        \includegraphics[width=\linewidth]{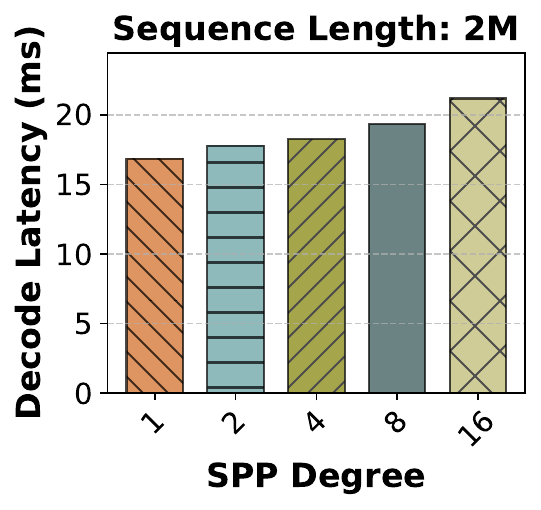}
        \caption{\llamaS}
        \label{fig:sppscaling:tbt:sevenb}
    \end{subfigure}
    \begin{subfigure}[b]{0.45\linewidth}
        \centering
        \includegraphics[width=\linewidth]{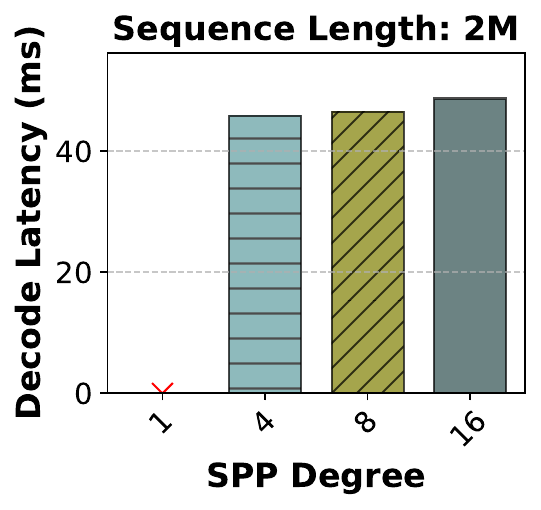}
        \caption{\llamaL}
        \label{fig:sppscaling:tbt:seventyb}
    \end{subfigure}
    \caption{
        Impact of SPP scaling on decode latency in \sysname{} 2D (SPP+TP, $p_{tp}=8$).
        Decode latency is only marginally affected even with a 16-stage pipeline.
    }
    \label{fig:sppscaling:tbt}
    \end{minipage}
    \hspace{0.5em}
    \begin{minipage}[t]{0.58\textwidth}
    \centering
    \begin{subfigure}[b]{0.49\linewidth}
        \includegraphics[width=\linewidth]{figures/experiments/cp_decode_scaling/cache_parallel_scaling_llama_7b.pdf}
        \caption{\llamaS with $p_{spp}=4$.}
        \label{fig:kvpscaling:decode:sevenb}
    \end{subfigure}
    \hfill
    \begin{subfigure}[b]{0.49\linewidth}
        \includegraphics[width=\linewidth]{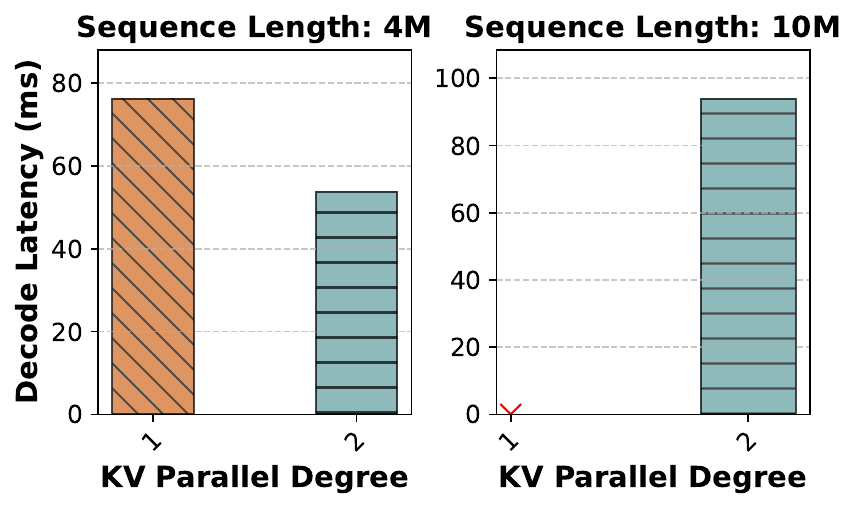}
        \caption{\llamaL with $p_{spp}=8$.}
        \label{fig:kvpscaling:decode:seventyb}
    \end{subfigure}
    \caption{
     TPOT reduction with KVP in \sysname 3D in decode-only batches.
     For 10M context length decodes for \llamaS, $p_{kvp} = 2$ results in almost 40\% reduction in latency, allowing decode at the rate of $\sim$30 tokens per second.
    }
    \label{fig:kvpscaling:decode}
    \end{minipage}
\end{figure*}

%% file: figures/experiments/cp_prefill_scaling/cp_scaling_prefill_tfft_all_fig.tex
\begin{figure*}
    \centering
    \includegraphics[width=0.9\textwidth]{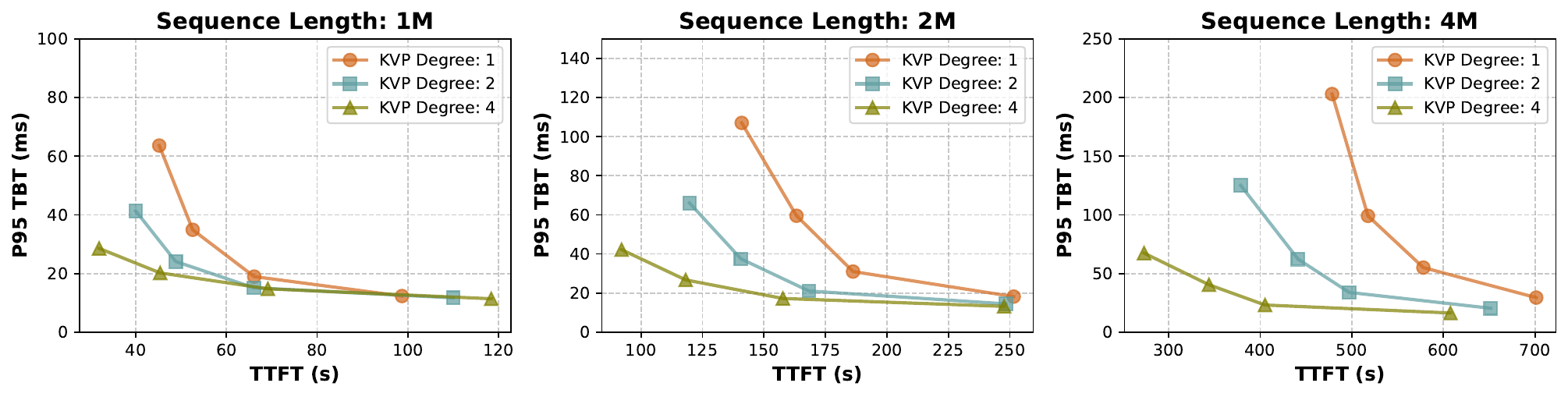}
    \caption{
    Trade-off Between TTFT and P95 TBT for \llamaS using \sysname 3D Parallelism ($p_{tp} = 4$, $p_{spp} = 4$) for varying KVP degrees and chunk sizes (32-256).
    }
    \label{fig:kvpscaling:prefill:ttft}
\end{figure*}